\pdfoutput=1
\documentclass[11pt]{article}

\usepackage[final]{acl}

\usepackage{times}
\usepackage{latexsym}

\usepackage[T1]{fontenc}
\usepackage[utf8]{inputenc}

\usepackage{microtype}

\usepackage{inconsolata}

\usepackage{graphicx}

\usepackage{epsfig}
\usepackage{graphicx}
\usepackage{amsmath,bm}
\usepackage{amssymb}
\usepackage{multirow}
\usepackage{verbatim}
\usepackage{enumitem}
\usepackage{booktabs,amsfonts,dcolumn}
\newcommand\blfootnote[1]{%
  \begingroup
  \renewcommand\thefootnote{}\footnote{#1}%
  \addtocounter{footnote}{-1}%
  \endgroup
}

\usepackage{xspace}

\makeatletter
\DeclareRobustCommand\onedot{\futurelet\@let@token\@onedot}
\def\@onedot{\ifx\@let@token.\else.\null\fi\xspace}

\def\eg{\emph{e.g}\onedot} 
\def\ie{\emph{i.e}\onedot}

\makeatother

\usepackage{amssymb}% http://ctan.org/pkg/amssymb
\usepackage{pifont}% http://ctan.org/pkg/pifont
\newcommand{\cmark}{\ding{51}}%
\newcommand{\xmark}{\ding{55}}%
\usepackage{multirow,multicol, makecell, booktabs}
\usepackage{array}
\usepackage{tcolorbox}

\newcommand{\vttab}{\newline\hspace*{1em}{}}
\newcommand{\vtred}[1]{\textcolor{red}{\textbf{#1}}}
\newcommand{\vtobj}[1]{\textcolor[RGB]{126, 171, 85}{\textbf{#1}}}
\newcommand{\vtattr}[1]{\textcolor[RGB]{222, 131, 68}{\textbf{#1}}}
\newcommand{\vtknow}[1]{\textcolor[RGB]{104, 52, 154}{\textbf{#1}}}
\usepackage{longtable}

\title{Beyond Embeddings: The Promise of Visual Table in Visual Reasoning}

\author{Yiwu Zhong$^{*1}$,
Zi-Yuan Hu$^{*12}$,
Michael R. Lyu$^1$, 
Liwei Wang$^{\dagger1}$ \\
$^1$The Chinese University of Hong Kong \quad 
$^2$Shanghai AI Laboratory \\
}

\begin{document}
\maketitle

\begin{abstract}
Visual representation learning has been a cornerstone in computer vision, involving typical forms such as visual embeddings, structural symbols, and text-based representations. 
Despite the success of CLIP-type visual embeddings, they often lack access to world knowledge critical for visual reasoning. 
In this work, we propose \textbf{Visual Table}, a novel form of visual representation tailored for visual reasoning.
Visual tables are constructed as hierarchical descriptions of visual scenes, featuring a scene description and multiple object-centric descriptions covering categories, attributes, and knowledge. 
Thanks to the structural and textual formats, visual tables offer \textbf{unique advantages} over mere visual embeddings, such as interpretability and controllable editing. Furthermore, they deliver instance-level world knowledge and detailed attributes that are essential for visual reasoning.
To create visual tables, we develop a generator trained on the dataset with collected, small-scale annotations.
Extensive results on \textbf{11} visual reasoning benchmarks demonstrate that the generated visual tables \textbf{significantly} outperform previous structural and text-based representations. Moreover, they \textbf{consistently} enhance state-of-the-art multimodal large language models across diverse benchmarks, showcasing their potential for advancing visual reasoning tasks. Our code is available at \url{https://github.com/LaVi-Lab/Visual-Table}. \blfootnote{$^*$equal contributions, $\dagger$ corresponding author.}
\end{abstract}

%-------------------------------------------------------------------------
% introduction
\section{Introduction}

Visual representation learning has been a fundamental and long-standing topic in computer vision. At the early stage, the learning was supervised by expensive human-annotated labels~\cite{NIPS2012_c399862d,szegedy2015going,he2016deep}. 
This paradigm recently evolved to learn visual embeddings by aligning image-text pairs from the Internet~\cite{radford2021learning,jia2021scaling}.  
Beyond visual embeddings, symbolic and structured visual representations (\eg, scene graph)~\cite{xu2017scene,krishna2017visual} exhibited advantages across domains, such as vision-language tasks~\cite{teney2017graph,hudson2019learning,zhong2020comprehensive}, video and 3D scene understanding~\cite{yang2023d,armeni20193d,wald2020learning}, and robotics~\cite{zhai2023sgbot,gu2023conceptgraphs,Kalithasan2023NSRM}. 
More recently, some works have strived to convert visual scenes into text-based representations (\eg, image captions)~\cite{hu2022promptcap,pica2022,prophet2023,mmreasoner2023}, triggering the reasoning capability of large language models (LLMs)~\cite{ouyang2022training, zhang2022opt, touvron2023llama, vicuna2023}.

\begin{figure*}[t]
    \centering
    \includegraphics[width=0.999\linewidth]{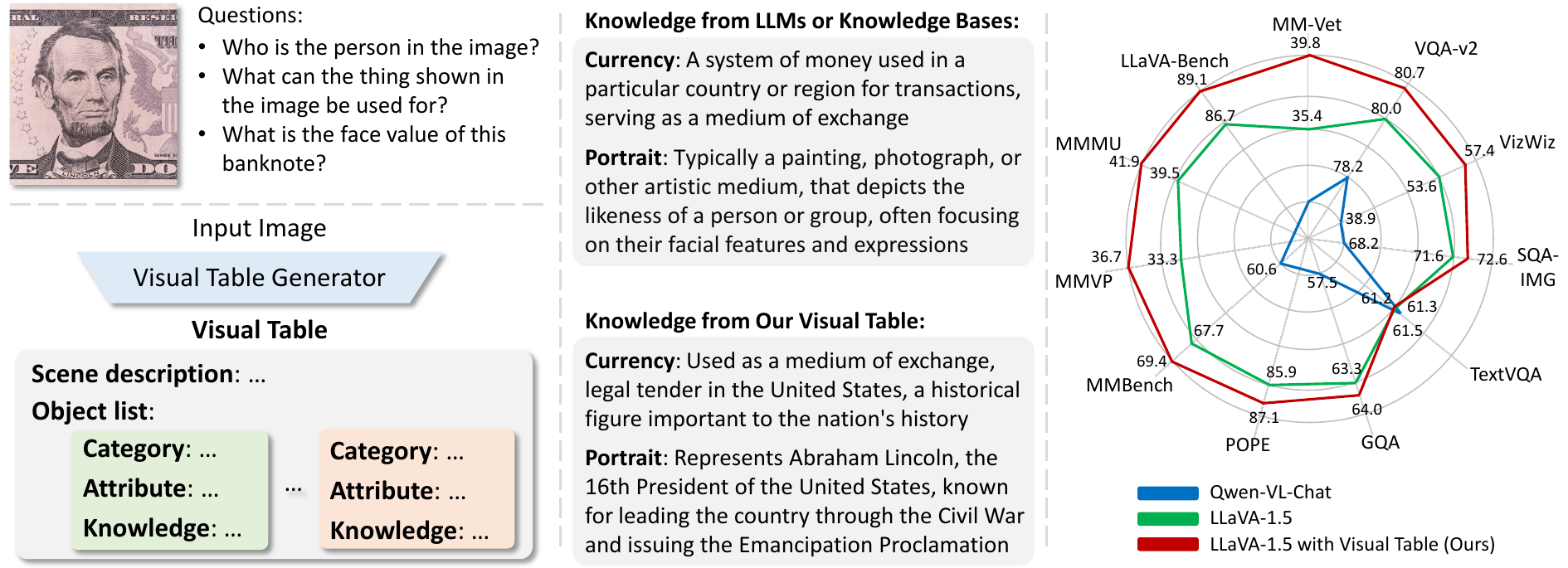}
    \vspace{-20pt}
    \caption{\textbf{Left Top:} An example that requires world knowledge to answer the questions about the image. \textbf{Left Bottom:} Our proposed visual tables represent thorough visual content in structured text. \textbf{Middle:} Visual tables provide instance-level knowledge in considering specific visual instances. \textbf{Right:} The generated visual tables can consistently enhance the state-of-the-art multimodal large language models across diverse benchmarks.}
    \vspace{-8pt}
    \label{fig:teaser}
\end{figure*}

Among these visual representations, CLIP-type visual embeddings~\cite{radford2021learning}, learned from image-text pairs, have dominated many vision tasks. Their success can be attributed to robust generalization in encoding visual attributes (\eg, visual appearance~\cite{yang2023language,pratt2023does,Yan_2023_ICCV}, visual relations~\cite{zhao2023unified,li2024zero,zhong2023learning,momeni2023verbs}). However, lacking external world knowledge (\eg, object affordance, background knowledge of named entities), these visual embeddings lead to sub-optimal performance in complex reasoning tasks. 
Consider the example at the left top of Fig.~\ref{fig:teaser}. While CLIP embeddings can effectively capture visual attributes (\eg, recognizing a person with a beard), they still struggle to answer questions that necessitate critical world knowledge beyond the image itself~\cite{marino2019ok,schwenk2022okvqa,chen2023can,yue2023mmmu}.

To this end, we propose \textbf{Visual Table}, a novel form of visual representation designed for visual reasoning. As shown in the left-bottom of Fig.~\ref{fig:teaser}, a visual table is presented as hierarchical text descriptions for visual scenes, and comprises a scene description and multiple object-centric descriptions, covering object categories, object attributes, and object knowledge. 
Through structural and textual formats, visual tables offer unique advantages beyond mere visual embeddings. 
First, \textbf{textual formats} can be seamlessly processed by LLMs without the need for training, while also facilitating interpretation and interaction with humans. 
Second, \textbf{hierarchical structures} support controllable table editing, enabling operations such as removal and insertion, and allowing customization of visual representation for different downstream tasks.
Third, by capturing visual scenes in \textbf{rich details}, such as the object-centric descriptions in free-form language covering attribute and knowledge, visual tables hold the potential to enhance visual reasoning across various domains, spanning images, videos, 3D scenes, and beyond. Note that \textbf{instance-level} object knowledge is generated in consideration of specific visual instances, unlike previous works that retrieve \textbf{category-level} knowledge from frozen LLMs or knowledge bases~\cite{pica2022,kat2021,revive2022,prophet2023}. For example, in the middle of Fig.~\ref{fig:teaser}, the visual table provides the world knowledge customized for the particular object instance, {\it a banknote of the United States with Abraham Lincoln who is the 16th President and leads the Civil War}, thereby supporting complex visual reasoning.

In this work, we focus on proof-of-concept for visual tables and validate its design and efficacy in image reasoning tasks.
To create our designed visual tables, a straightforward solution is to collect human annotations and use them to train a generator.
Despite high quality, it is costly in terms of the training required for human annotators and the label normalization during post-processing. To reduce these burdens, we adopt an alternative solution leveraging foundation models with near-human capabilities~\cite{GPT4v,gemini,Claude3}. Specifically, we design prompts to ensure the annotations closely align with our specifications.
After collecting a small scale of annotations, \ie, 61K visual tables, we train a generator capable of automatically producing visual tables for arbitrary images.

To validate the effectiveness of visual tables as visual representations, we adopt \textbf{11 diverse benchmarks} that require an understanding of object attributes and knowledge. 
We seek to answer two key questions: (1) Do visual tables represent visual scenes more effectively than traditional structural and text-based representations (\eg, scene graphs, image captions)? (2) Can visual tables enhance the performance of existing multimodal large language models (MLLMs), which already demonstrated incredible results on visual reasoning~\cite{dai2023instructblip, liu2023improved, bai2023qwen}?
Our extensive experiments reveal that visual tables \textbf{significantly} outperform previous structural and text-based representations across various benchmarks. Furthermore, visual tables \textbf{consistently} improve the performance of state-of-the-art (SOTA) MLLMs (\eg, LLaVA-1.5~\cite{liu2023improved}), as shown at the right of Fig.~\ref{fig:teaser}. These findings validate that visual tables can facilitate visual reasoning by serving as general, robust visual representations.

Our contributions are summarized as follows: 
(1) We propose Visual Table, a new form of visual representation organized in structural text. It offers unique benefits beyond visual embeddings --- interpretability, controllable editing, and instance-level knowledge.
(2) We introduce a new dataset with 61K visual table annotations, and present a generator that can produce high-quality visual tables for any input images. The dataset, together with our visual table generator, can be further exploited by future research.
(3) Extensive experiments show that visual tables, working as generalizable representations, largely outperform previous structural, text-based representations and consistently improve the SOTA MLLMs across benchmarks.

%-------------------------------------------------------------------------
% related works
\section{Related Work}

\noindent\textbf{Visual Representation Learning.} Early works focus on training image classifiers using labor-intensive image labels~\cite{NIPS2012_c399862d,simonyan2014very,szegedy2015going,he2016deep}. 
As an augmentation, object and attribute labels are used to enhance visual representation~\cite{Anderson2017up-down,Zhang_2021_CVPR}.
To reduce the annotation cost, self-supervised learning~\cite{he2020momentum,chen2020simple,grill2020bootstrap,He_2022_CVPR} is proposed to match the visual representation of different views from the same image. Moving forward, vision-language pre-training~\cite{radford2021learning,jia2021scaling,gu2021zero,zhong2022regionclip,li2021grounded} is proposed to match web-collected visual-text pairs, exhibiting generalizable capability on diverse visual recognition tasks. Compared to these visual embeddings, our visual tables additionally encode \textbf{world knowledge} critical for visual reasoning, support \textbf{interpretability} to humans and LLMs, and enable \textbf{controllable editing}.

Beyond implicit embeddings, structural and symbolic representations, such as scene graphs~\cite{xu2017scene,zellers2018neural,tang2020unbiased,zhong2021learning}, have attracted significant attention. These works aim to abstract visual scenes into concise representations, demonstrating special benefits in diverse domains, including vision-language modeling~\cite{yu2021ernie,pan2022contrastive,mitra2023compositional,herzig2023incorporating} and various downstream tasks~\cite{hudson2019learning, zhong2020comprehensive, ji2020action, hughes2022hydra, Kalithasan2023NSRM}. Resembling the concept of structural representations, visual tables are presented in the hierarchical text, yet deliver \textbf{richer semantics} through free-form language.

Another line of works explores text-based visual representation~\cite{hu2022promptcap,filling2023, pica2022,kat2021,revive2022,prophet2023,RASO2023, mmreasoner2023, VidIL2022, hakimov2023}. These methods typically convert visual inputs into text (\eg, image captions, object tags), then retrieve knowledge from knowledge bases (\eg, Wikipedia~\cite{wikipedia2014}, ConceptNet~\cite{conceptnet2004}) and/or frozen LLMs (\eg, GPT3), and finally perform text reasoning using frozen LLMs with in-context examples. 
Unlike these works, our \textbf{single} generator model learns to compress a comprehensive visual knowledge base, \textbf{without the need} for manually selecting off-the-shelf vision APIs (\eg, image captioners, object detectors), external knowledge sources, or high-quality in-context examples.

\begin{figure*}[t]
    \centering
    \includegraphics[width=0.999\linewidth]{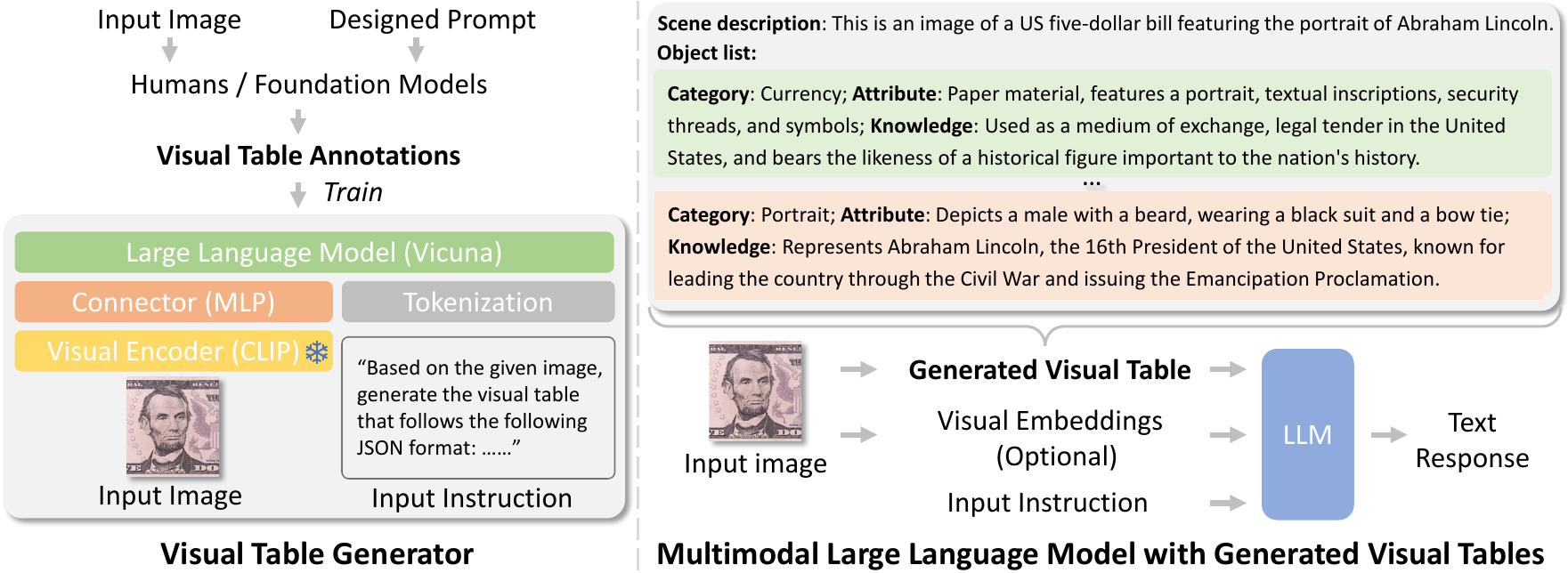}
    \vspace{-20pt}
    \caption{An overview of learning a visual table generator and its application on MLLMs. \textbf{Left:} We design a prompt to collect visual table annotations on a small scale of images. These annotations are used to train our visual table generator which consists of a frozen visual encoder, a vision-language connector, and a pre-trained LLM. \textbf{Right:} Our generator is employed to generate visual tables for the images in downstream tasks. With the generated visual table either as standalone or as additional visual representations, LLM performs reasoning to output text response.}
    \vspace{-8pt}
    \label{fig:method}
\end{figure*}

\smallskip
\noindent\textbf{Multi-modal Large Language Models.} MLLMs harness LLMs to empower reasoning on multi-modal tasks, typically on visual question answering (VQA)~\cite{li2023seed,MME,yin2023survey,alayrac2022flamingo, dai2023instructblip, zhu2023minigpt4, chen2023shikra, bai2023qwen, idefics2023, huang2024language, li2023otter}. These methods usually learn layers that connect visual encoder~\cite{radford2021learning} and LLMs~\cite{touvron2023llama, vicuna2023}. 
Building on top of MLLMs, some works seek to improve the quality of instruction-following data~\cite{chen2023sharegpt4v,wang2023see} or to enhance object perception by introducing control signals~\cite{jain2023vcoder}, while the others explore chain-of-thoughts idea~\cite{kojima2022large,wei2022chain,  mmcot2023,ddcot2024,  mitra2023compositional}. 
In parallel, we focus on visual representation learning and the resulting visual tables serve as inputs of MLLMs and above methods.

%-------------------------------------------------------------------------
% method
\section{Visual Table}

We introduce Visual Table, a new form of visual representation, constructed in hierarchical text. Given an image $\mathbf{I}$, a visual table $\mathbf{V}_{\rm T} = g(\mathbf{I})$ is created by a generator $g$. $\mathbf{V}_{\rm T}$ consists of a scene description and multiple object-centric descriptions that encompass categories, attributes, and knowledge. 
Thanks to structural and textual formats, visual tables support interpretability to humans and LLMs, and enable controllable table editing. Moreover, they capture visual scenes at a level of granularity (\eg, instance-level knowledge) and richness (\eg, free-form language). These characteristics render visual tables potentially beneficial to a broad spectrum of vision tasks.

Fig.~\ref{fig:method} provides an overview of visual table generation and its application on image reasoning using MLLMs. We first collect a small number of visual table annotations (Sec.~\ref{Training Data Collection}) and then train a visual table generator $g$ (Sec.~\ref{Visual Table Generation}). The trained generator is able to automatically produce visual tables for any input images in downstream tasks. We evaluate the efficacy of our generated visual tables on diverse benchmarks where visual tables serve as visual representations (Sec.~\ref{MLLMs with Visual Table}).

\subsection{Training Data Collection}
\label{Training Data Collection}
To create visual table annotations, a straightforward solution is to ask human annotators to label images with the required visual table content. Despite high quality, training human annotators and cleaning their annotations can be expensive and cumbersome. To mitigate these burdens, we opt for an alternative solution by leveraging powerful foundation models with near-human capabilities, such as GPT4V \cite{GPT4v}. We design a detailed prompt to ensure the annotations closely align with our requirements, resulting in a small scale of visual tables, \ie, on 61K images~\cite{lin2014microsoft}.

\smallskip
\noindent \textbf {Prompt design.} 
We design a prompt consisting of performing four tasks on input images: scene description generation, object category recognition, attribute description generation, and knowledge description generation. Without losing generality, we provide \textbf{basic definitions and guidelines} for each task to ensure versatile descriptions. For example, we define object affordance as {\it the functions supported by objects} and remind that {\it affordance might be altered case by case, due to deformed shape, unreliable materials, and so on}.

At the end of the prompt, we specify that the output should be in JSON format, \ie, a nested dictionary. This structure design enables \textbf{controllable manipulation} on visual tables, such as studying the effects of removing table components. The full prompt and the collected visual tables can be found in the appendix.

\smallskip
\noindent \textbf {Statistics.} We collect a visual table per image, in total 61K images from the COCO dataset~\cite{lin2014microsoft}. These images are selected based on the scheme that an image should be associated with at least two instruction-following responses in the LLaVA-Instruct-158K dataset~\cite{liu2023improved}. On average, there are 458 text tokens in each visual table after the tokenization process.

\subsection{Visual Table Generation}
\label{Visual Table Generation}

With annotations ready, we train a model to generate visual tables. Learning such a generator is challenging and requires two key capabilities: (1) robust visual perception and (2) reliable text generation. 
To this end, we leverage existing MLLMs due to their remarkable performance in both visual perception and text generation.

\smallskip
\noindent \textbf {Visual Table Generator.} 
Our generator follows the architecture of typical MLLMs. It consists of a visual encoder that converts an input image into visual embeddings, a connector that connects visual embeddings and the LLM, and a pre-trained LLM that performs reasoning and outputs text responses. 
The model is trained to predict the next token in an auto-regressive way:
\begin{equation}
\resizebox{.43\textwidth}{!}{
$p( \mathbf{T}_{\rm a} | \mathbf{I}, \mathbf{T}_{\rm instruct}) = \prod_{i}^{L} p_{\bm{\theta}}(\mathbf{t}_i | h(\mathbf{I}), \mathbf{T}_{{\rm instruct},<i}, \mathbf{T}_{{\rm a},<i}),$
}
\label{auto-regressive}
\end{equation}
where $L$ denotes sequence length, $\bm{\theta}$ denotes trainable parameters (\eg, connector and LLM), $h$ represents frozen visual encoder, $\mathbf{T}_{{\rm instruct},<i}$ and $\mathbf{T}_{{\rm a},<i}$ are text tokens of instructions and answers before the current prediction token $\mathbf{t}_i$, respectively. 

\smallskip
\noindent \textbf {Training and Inference.} Our generator is trained on the collected visual table annotations and partial training data from LLaVA-1.5~\cite{liu2023improved}. Specifically, there are three training stages: (1) Visual-language alignment: With visual encoder and LLM frozen, connector is trained on 595K image-text pairs by instructing the model to generate captions. (2) Instruction fine-tuning: With visual encoder frozen, connector and LLM are trained on 199K GPT-generated instruction-tuning data. (3) Supervised fine-tuning: With visual encoder frozen, connector and LLM are trained on our 61K visual table annotations. Once trained, our generator can automatically generate a visual table as the visual representation of any given image.

Note that, we avoid using any human annotations from visual question answering (VQA) datasets during training, \ie, 467K VQA instances that were used to optimize VQA performance of MLLMs. This strategy aims to \textbf{minimize the biases} from VQA tasks, thereby allowing the resulting visual tables to potentially benefit a broader spectrum of visual tasks beyond VQA.

\smallskip
\noindent \textbf {Statistics.} Our generator is used to produce visual tables for all images in evaluation benchmarks, including training and test images. The average number of tokens is 421 after tokenization process.

\subsection{Application on Image Reasoning}
\label{MLLMs with Visual Table}
With instance-level descriptions of attributes and knowledge, visual tables essentially create comprehensive databases for individual visual scenes. 
In this work, we focus on its application in image reasoning tasks and assess its impact on performance within the context of MLLMs.

Specifically, visual tables are first generated for the images. We then re-train the MLLMs according to their original training methodologies, using the generated visual tables as visual representations. 
We choose the LLaVA-1.5 as our main testbed. The model training follows its original pipeline, except that at its second training stage (instruction-tuning stage), the LLM module takes the generated visual tables $\mathbf{V}_{\rm T} = g(\mathbf{I})$ as the input, and learns to predict next token:
\begin{equation}
\resizebox{.43\textwidth}{!}{
$p( \mathbf{T}_{\rm a} | \mathbf{I}, \mathbf{T}_{\rm instruct}) = \prod_{i}^{L} p_{\bm{\theta}}(\mathbf{t}_i | g(\mathbf{I}), h(\mathbf{I}), \mathbf{T}_{{\rm instruct},<i}, \mathbf{T}_{{\rm a},<i}).$
}
\end{equation}
We highlight that our visual tables $g(\mathbf{I})$ are text formats and thus can function as standalone visual representations, without using visual embeddings $h(\mathbf{I})$. In this scenario, the generated visual tables can be \textbf{directly} processed by pre-trained LLMs \textbf{without} requiring the first training stage for visual-language embedding alignment. Our experiments demonstrate that using only visual tables already achieves strong performance across diverse benchmarks, while combining both visual representations leads to further performance improvements.

%-------------------------------------------------------------------------
% experiments
\section{Experiments}
In this section, we first introduce our implementation details, benchmarks, and evaluation protocols, and then present our results. 
%including the comparison with traditional text-based representations, the comparison with SOTA MLLMs, the ablation study, and the case study.  

\begin{table*}[tb]
  \vspace{-8pt}
  \centering
  \resizebox{\textwidth}{!}{
  \begin{tabular}{l lcc | c | c | c | c | c | c | c | c | c | c | c }
    \toprule
    Method & LLM & \#PT & \#IT & MM-Vet & LLaVA$^\text{W}$ & MMMU & MMB & MMVP  & POPE & VizWiz & SQA$^\text{I}$ & GQA & VQA$^\text{v2}$ & VQA$^\text{T}$  \\
    \midrule
    \midrule
    \textbf{Representation: E} &  &  &  &  &   &  &  &  &  &  & &  & & \\
    BLIP-2~\cite{li2023blip} & V-13B   & 129M & - & 22.4 & -- & --  & -- & -- & 85.3 & 19.6 & 61.0 & 41.0 & 41.0 & 42.5  \\
    InstructBLIP~\cite{dai2023instructblip} & V-7B   & 129M & 1.2M & 26.2 & -- & -- & 36.0 & -- & -- & 34.5 & 60.5 & 49.2  & -- & 50.1  \\
    InstructBLIP~\cite{dai2023instructblip} & V-13B   & 129M & 1.2M & 25.6 & -- & -- & -- & -- & 78.9 & 33.4 & 63.1 & 49.5 & -- & 50.7  \\
    Shikra~\cite{chen2023shikra} & V-13B   & 600K & 5.5M & -- & --  & -- & 58.8 & -- & -- & -- & -- & -- & 77.4 & -- \\
    IDEFICS-9B~\cite{idefics2023} & L-7B  & 353M & 1M & -- & --  & -- & 48.2 & -- & -- & 35.5 & -- & 38.4 & 50.9 & 25.9  \\
    IDEFICS-80B~\cite{idefics2023} & L-65B & 353M & 1M & -- & -- & -- & 54.5 & -- & -- & 36.0 & -- & 45.2 & 60.0 & 30.9 \\
    Qwen-VL~\cite{bai2023qwen} & Q-7B  & 1.4B  & 50M  & -- & -- & -- & 38.2 & -- & -- & 35.2 & 67.1 & 59.3 & 78.8 & \textbf{63.8} \\
    Qwen-VL-Chat~\cite{bai2023qwen} & Q-7B   & 1.4B   & 50M  & -- & -- & -- & 60.6 & -- & -- & 38.9 & 68.2 & 57.5 & 78.2 & {61.5} \\
    LLaVA-1.5~\cite{liu2023improved} & V-7B  & 558K & 665K & 30.5 & 81.9 & 33.9 & 64.3 & 20.7 & \textbf{85.9} & 50.0 & 66.8 & 62.0 & 78.5 & 58.2 \\
    LLaVA-1.5~\cite{liu2023improved} & V-13B & 558K & 665K & \textbf{35.4} & \textbf{86.7} & \textbf{39.5} & \textbf{67.7} & \textbf{33.3} & \textbf{85.9}  & \textbf{53.6} & \textbf{71.6} & \textbf{63.3} & \textbf{80.0} & 61.3 \\
    \midrule
    \textbf{Representation: T} &   &  &  &  &   &  &  &  &  &  & &  & & \\
    Vicuna-Cap & V-13B  & -- & 665K & 23.0 & 79.2 & 39.1 & 62.1 & 12.0 & 73.3 & 51.3 & 69.5 & 48.4 & 61.4 & 48.0 \\
    Vicuna-DCap & V-13B  & -- & 665K & 27.1 & 77.6 & 37.4 & 61.4 & 13.3 & \textbf{83.4}  & 51.6 & 69.3 & 51.7 & 68.5 & 49.6 \\
    Vicuna-SG & V-13B  & -- & 665K & 28.1 & 77.0 & 36.5 & 59.3 & 11.3 & 82.3 & 51.0 & 68.9 & 52.0 & 67.8 & 49.3 \\
    \textbf{Vicuna-VT} & V-13B  & -- & 665K & \textbf{30.7} & \textbf{82.5} & \textbf{39.6} & \textbf{62.7} & \textbf{26.7} & 81.9 & \textbf{55.4} & \textbf{70.0} & \textbf{56.1} & \textbf{74.0} & \textbf{53.8}  \\
    \midrule
    \textbf{Representation: E + T} &  &  &  &  &   &  &  &  &  &  & &  & & \\
    LLaVA-Cap & V-13B  & 558K & 665K & 36.3 & {88.9} & {40.6} & \textbf{69.4} & 32.0 & 86.4 & 53.8 & {71.9} & {63.5} & {80.3} & 60.7  \\
    LLaVA-DCap & V-13B  & 558K & 665K & {36.7} & 86.4 & 38.6 & 68.9 & 30.7 & 86.9 & 52.6 & 71.2 & 63.4 & {80.3} & 60.7 \\
    LLaVA-SG & V-13B  & 558K & 665K & 36.1 & 86.8 & {40.6} & {69.2} & 30.0  & \textbf{87.3} & \textbf{57.5} & 71.8 & 63.1 & {80.3} & 61.1 \\
    \textbf{LLaVA-VT} & V-13B  & 558K & 665K & \textbf{39.8} & \textbf{89.1} & \textbf{41.9} & \textbf{69.4} & \textbf{36.7} & {87.1} & {57.4} & \textbf{72.6} & \textbf{64.0} & \textbf{80.7} &  \textbf{61.2} \\
    \bottomrule
  \end{tabular}
  }
  \vspace{-3pt}
  \caption{\textbf{Comparison with text-based representations and MLLMs.}  
  \textbf{E}, \textbf{T} and \textbf{E + T} denotes the visual representations as visual embeddings, text-based representations ({Cap}: Short Caption; {Dcap}: Detailed Caption; {SG}: Scene Graph; {VT}: Visual Table) and their concatenation, respectively.
  \#PT/\#IT denotes the number of samples in the stage-one/two training, respectively.
  V-7B/13B: Vicuna-7B/13B~\cite{vicuna2023}; L-7B/13B: LLaMA-7B/13B~\cite{touvron2023llama}; Q-7B/13B: Qwen-7B/13B~\cite{bai2023qwen}. 
  {Bold} values refer to the best results within each group. 
  Visual table \textbf{largely outperforms} previous text-based representations and is the only text representation that can \textbf{consistently} enhance SOTA MLLMs across diverse benchmarks.
  }
  \label{tab:main}
\end{table*}

\smallskip
\noindent\textbf{Implementation Details.} 
%We provide implementation details for our visual table generator and the MLLM with visual tables. 
For \textbf{visual table generator}, we adopt the same architecture of LLaVA-1.5~\cite{liu2023improved}, consisting of CLIP ViT-L/14@336px~\cite{radford2021learning} as the visual encoder, Vicuna-13B~\cite{vicuna2023} as the LLM, and a two-layer MLP as the connector. It is initialized by the LLaVA-1.5-13B model pre-trained by excluding VQA data, and further fine-tuned on 61K visual table annotations for 3 epochs, with batch size as 128,  learning rate as 2e-5, and the optimizer as AdamW.
For the \textbf{MLLM with visual table}, we adopt the same training pipeline as LLaVA-1.5, except that during the second training stage, our model is fine-tuned with the generated visual tables as additional visual representations.

\smallskip
\noindent\textbf{Benchmarks.}
To evaluate visual tables, we conduct experiments across a diverse set of 11 evaluation benchmarks, providing a comprehensive assessment of visual reasoning capability.
Our evaluation set encompasses both recent benchmarks designed for MLLMs, including MM-Vet~\cite{yu2023mm}, LLaVA-Bench~\cite{llava}, MMMU~\cite{yue2023mmmu}, MMBench~\cite{liu2023mmbench}, MMVP~\cite{tong2024mmvp} and POPE~\cite{li2023pope}, and academic VQA benchmarks, including VizWiz~\cite{gurari2018vizwiz}, ScienceQA~\cite{lu2022sqa}, GQA~\cite{hudson2019gqa}, VQA-v2~\cite{goyal2017vqav2}, and TextVQA~\cite{singh2019textvqa}.

\smallskip
\noindent\textbf{Evaluation Protocols.}
We adopt two widely-used protocols: (1) Exact-matching protocol matches the predicted answer string and ground-truth string~\cite{goyal2017vqav2,hudson2019gqa}. (2) GPT-assisted protocol relies on GPT models to measure the correctness of the predicted, open-ended answer, given the question and ground-truth answer~\cite{yu2023mm,llava}.

\subsection{Comparison Experiments}
\label{Comparison with Text-based Representations}
Tab.~\ref{tab:main} shows the results of typical text-based representations, our visual tables, and recent MLLMs.

\smallskip
\noindent\textbf{Setup.} The exact-matching evaluation is utilized for academic VQA benchmarks, including POPE, VizWiz, ScienceQA, VQA-v2, GQA, and TextVQA. 
This evaluation protocol is also applied to MMBench and MMVP benchmarks due to their multiple-choice settings. 
We use GPT-assisted evaluation for the remaining benchmarks, including MM-Vet (open-ended VQA, using the official GPT-4 evaluation server), LLaVA-Bench (open-ended VQA, using GPT-3.5-1106), and MMMU (open-ended VQA, with 855 VQA paris sampled from its original val split, using GPT-3.5-1106).

\smallskip
\noindent\textbf{Baselines.}
Visual table is designed as a structural text-based representation. We thus compare it with the commonly-used text-based representations, including:
(1) \textbf{Cap}: We generate short captions using BLIP2-OPT-2.7B~\cite{li2023blip}, a captioner from the widely-used BLIP model family. LLaVA-1.5 also employs a captioner from this family to create its training data. 
(2) \textbf{DCap}: We use the same pre-trained LLaVA-1.5 model as our generator to produce detailed captions since it already learned image captioning during training. The difference is that our generator is further fine-tuned on visual table annotations.
(3) \textbf{SG}: Inspired by~\cite{mitra2023compositional}, we utilize the same pre-trained LLaVA-1.5 model as our generator to produce scene graphs, including visual attributes and visual relationships. Both scene graphs and visual tables are structural representations, while visual tables have richer semantic descriptions and cover object knowledge.

With these text-based representations ready, we replace visual tables and re-train MLLMs. When used as standalone visual representation, without the need for visual embeddings or the need for visual-language alignment training, they are denoted as \textbf{Vicuna-Cap/DCap/SG}, respectively.  
When combined with visual embeddings, they are denoted as \textbf{LLaVA-Cap/DCap/SG}, respectively. 

Further, to validate whether visual tables can enhance existing MLLMs that have already exhibited incredible results on visual reasoning, we incorporate baselines of recent MLLMs, \eg, BLIP-2~\cite{li2023blip}, InstructBLIP~\cite{dai2023instructblip}, Shikra~\cite{chen2023shikra}, IDEFICS~\cite{idefics2023}, Qwen-VL~\cite{bai2023qwen} and LLaVA-1.5~\cite{liu2023improved}.

\smallskip
\noindent\textbf{Comparison with Text-based Representations.} 
As Tab.~\ref{tab:main} shows, no matter in Vicuna-VT setting or LLaVA-VT setting, visual tables \textbf{significantly} outperform traditional text-based representations (\eg, +2.6 on MM-Vet, +5.5 on LLaVA-Bench, +3.1 on MMMU, +15.4 on MMVP over Vicuna-SG).
Moreover, it is worth noting that visual table (LLaVA-VT) is the \textbf{only} text-based representation that can \textbf{consistently improve} the base model LLaVA-1.5 across benchmarks.
These results reveal that, unlike previous text-based representations, visual tables stand out as robust visual representations capable of generalization across various scenarios.

\smallskip
\noindent\textbf{Consistent Improvements over SOTA MLLMs.}
Despite the challenge of further enhancing existing MLLMs, which have already demonstrated remarkable performance in visual reasoning, visual tables (LLaVA-VT) \textbf{consistently} achieve improvements over SOTA MLLMs across diverse benchmarks (\eg, +4.4 on MM-Vet, +2.4 on LLaVA-Bench, +2.4 on MMMU, +3.4 on MMVP). These consistent enhancements across various benchmarks validate the \textbf{generality} of our visual table design and can be attributed to the valuable information provided by visual tables, such as world knowledge, which visual embeddings lack.

\smallskip
\noindent\textbf{Textual Representations vs.~Visual Embeddings.} It is not surprising that visual embeddings perform better than text-based representations (LLaVA-1.5 vs.~Vicuna-Cap/DCap/SG/VT). 
However, it is worth recognizing that text representations by nature offer \textbf{unique advantages} over visual embeddings. 
For instance, textual formats can be directly utilized by LLMs without the need for visual-language alignment (\eg, Vicuna-VT), offering a solution for domains where alignment data is scarce, such as 3D vision.
Textual formats are also interpretable to humans, facilitating applications that require human-machine interaction. 
Moreover, by explicitly representing object knowledge in text, visual tables can support research such as embodied AI that requires an understanding of object affordance~\cite{gu2023conceptgraphs}.
Overall, we believe that text-based representations are worth exploring and our work suggests a promising avenue of representing visual scenes in structural and rich text.

\begin{table}[tb]
% \vspace{-5pt}
\centering
\setlength{\tabcolsep}{8pt}
\resizebox{0.49\textwidth}{!}{
    \begin{tabular}{ccc|c|c|c|c}
    \toprule
    \multicolumn{3}{c|}{\textbf{Visual Table}} & \multicolumn{1}{c|}{\textbf{GQA}} & \multicolumn{1}{c|}{\textbf{MMVP}} & \multicolumn{1}{c|}{\textbf{MM-Vet}} & \multicolumn{1}{c}{\textbf{MMMU}} \\  
    %\cmidrule{0-8}
      \textbf{Scene} & \textbf{Attribute} & \textbf{Knowledge} & \textbf{Accuracy}  & \textbf{Accuracy} & \textbf{Accuracy} & \textbf{Accuracy} \\
    \midrule
\midrule
\xmark & \xmark &\xmark & 56.7 & 18.0 & 35.6 & 42.3 \\ 
\cmark & \xmark &\xmark & 56.8 & 20.0 & 41.2 & 41.3 \\
\cmark& \cmark & \xmark& \textbf{58.5}  & 20.0 & 41.8 & 44.0   \\
\cmark& \xmark & \cmark& 56.3  & \underline{20.7} & \textbf{43.7} & \textbf{45.4}  \\
\cmark& \cmark & \cmark& \underline{57.8} & \textbf{21.3} & \underline{43.0} & \underline{44.8}   \\
    \bottomrule
    \end{tabular}
    }
    \vspace{-5pt}
    \caption{\textbf{Ablation study on visual table components}. We probe the components by editing visual tables and re-training our MLLM. Bold/underlined values refer to the best/second-best results.}
    \label{tab:vt_ablation}
    \vspace{-5pt}
\end{table}

\subsection{Ablation Study}
\label{Ablation Study}

Benefiting from the structural design, the components in visual tables can be effortlessly edited. This enables ablation experiments to probe the impacts of removing certain components in visual tables. We use the edited visual tables as visual representations and re-train MLLMs for comparisons.  

\smallskip
\noindent\textbf{Setup.} 
In the ablation study, we utilize 61K annotated visual tables to inspect the impacts of each component since they follow strict JSON formats. 
All MLLMs are first trained on 595K image-text pairs, and then fine-tuned on 177K instruction-following data with the edited visual tables as inputs.  
GPT-assisted evaluation is used for benchmarks: GQA (open-ended VQA with 398 VQA pairs sampled from test-dev split, using GPT-3.5-1106), MMVP (open-ended VQA, using the GPT-3.5-1106), MM-Vet (open-ended VQA, using their official GPT-4 evaluation server), and MMMU (open-ended VQA with 855 VQA paris sampled from its original val split, using GPT-3.5-1106).

\smallskip
\noindent\textbf{Results.} Tab.~\ref{tab:vt_ablation} presents the results of our ablation study. 
Compared to the baseline model (row 1) that takes CLIP visual embeddings as the only visual representations, scene descriptions (row 2) improve the performance on MMVP (+2.0) and MM-Vet (+5.6), yet bring limited benefits on GQA (+0.1) and worse performance on MMMU (-1.0). These results suggest that scene descriptions can provide useful information but \textbf{cannot robustly} benefit wide benchmarks. 
When compared with row 2, adding attributes (row 3) largely improves the performance on GQA (+1.7), and adding knowledge (row 4) significantly improves the results on MM-Vet (+2.5) and MMMU (+4.1). These results align with intuition since GQA highlights object \textbf{attributes} while MM-Vet and MMMU heavily rely on \textbf{knowledge} to answer the questions.

Combining all components, full visual tables (row 5) achieve either the best or the second-best results across all benchmarks, striking a good balance. Notably, full visual tables (row 5) largely outperforms scene descriptions (row 2), even though both are annotated by GPT4V. These results validate the necessity of \textbf{all} components and the consistent performance improvements stem from \textbf{our design} of visual tables, instead of the annotation tool.

\begin{figure*}[t] % [p]
    \centering
    \includegraphics[width=1.0\linewidth]{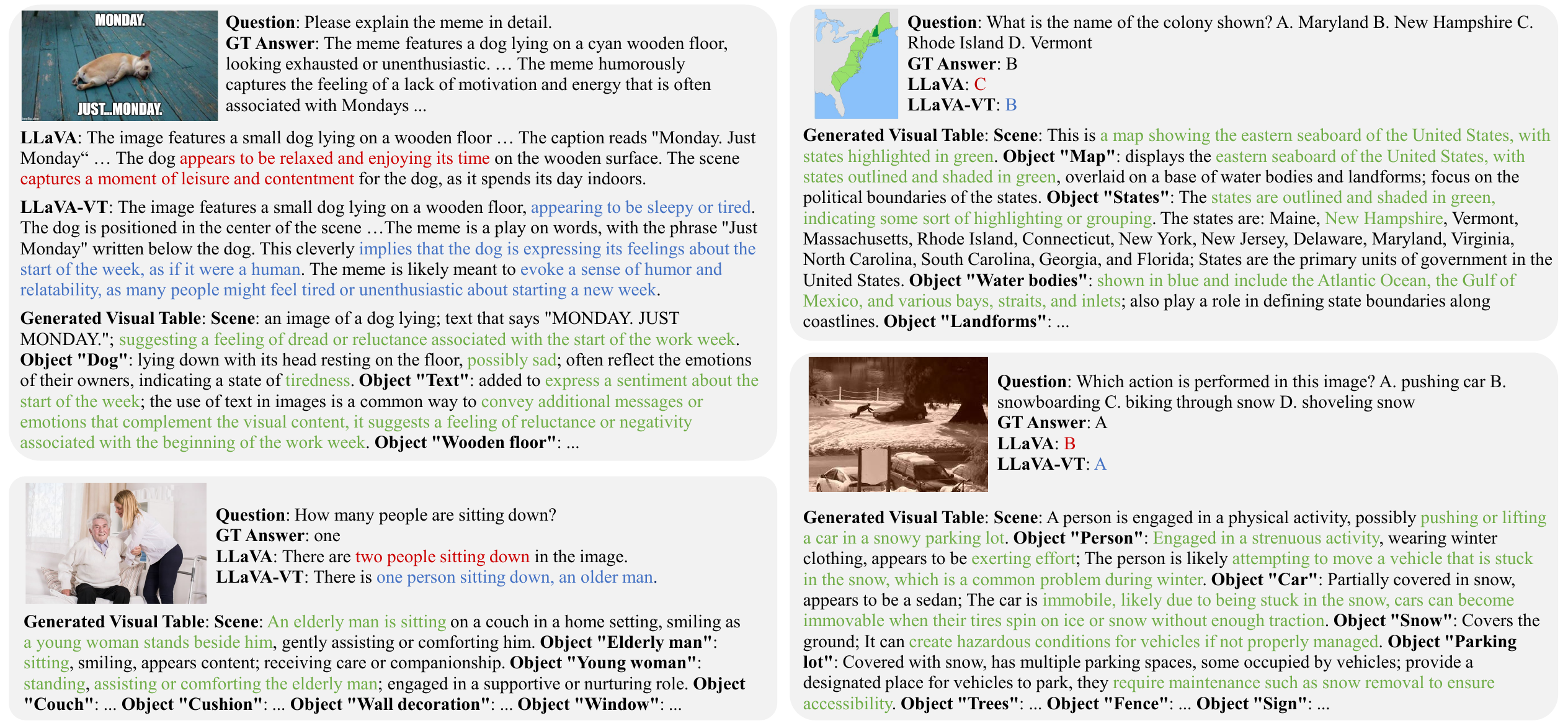}
    \vspace{-20pt}
    \caption{Visualization of visual reasoning examples. 
    For simplicity, we visualize partial visual tables that relate to the question, with attribute and knowledge separated by ``;''. Best viewed in color.}
    \label{fig:case_subset}
    \vspace{-5pt}
\end{figure*}

\subsection{Visualization and Case Study}
\label{case_discussion}

Benefiting from the textual formats, visual tables allow humans to interpret and inspect how they support visual reasoning, as illustrated in Fig.~\ref{fig:case_subset}.
For simplicity, we show partial visual tables, \ie, the evidence that supports the visual question answering. We highlight the wrong answers in \textbf{red}, the correct answers in \textbf{blue}, and the supporting evidence from visual tables in \textbf{green}, respectively. 
More examples can be found in the appendix.

Take the top-left image as an example. Our model can correctly describe the theme of the image as ``sleepy or tired'', instead of the answer ``relaxed and enjoying its time'' with opposite semantics from the LLaVA model. Our correct answer is attributed to the generated visual table. It identifies the dog as ``possibly sad'', provides \textbf{the knowledge of the dog instance} as ``often reflect the emotions of their owners, indicating a state of tiredness'', and explicitly offers the \textbf{world knowledge} ``the beginning of the work week'' based on the text ``MONDAY. JUST MONDAY''. 
Moreover, visual tables can provide \textbf{discipline knowledge}, such as geographic knowledge in the top-right example, ``the eastern seaboard of the United States'' and ``The states are: Maine, New Hampshire, ...''.

Besides knowledge, visual tables can \textbf{precisely recognize object attributes}. For instance, ``Elderly man: sitting; Young woman: standing, assisting or comforting the elderly man'' is identified in the bottom-left example, thereby facilitating the correct answer. 
Further, visual tables explicitly record the perceived visual objects and their knowledge, \textbf{reducing the chance of hallucinations} during LLM reasoning. Consider the bottom-right image. ``Person: attempting to move a vehicle that is stuck in the snow, which is a common problem during winter'' in the visual table can avoid the wrong answer of ``snowboarding'' which also often happens during snowy days.

In summary, while being interpretable to humans, visual tables thoroughly describe the detailed objects in visual scenes and provide precise attributes and rich knowledge, thereby consistently improving SOTA MLLMs across diverse benchmarks, as demonstrated in experiments.

\section{Conclusion}
In this paper, we propose \textbf{Visual Table} --- a new visual representation presented in structured text.
Using the collected dataset with visual table annotations, we learn a generator to produce high-quality visual tables for arbitrary input images. 
Beyond visual embeddings, visual tables support interpretability to humans and LLMs, enable controllable editing, and more importantly, offer rich descriptions of attributes and knowledge for each object, thereby facilitating visual reasoning.
According to extensive experiments, the resulting visual tables exhibit superior performance than previous text-based representations, and demonstrate consistent improvements over the SOTA MLLMs across diverse benchmarks.
We believe our study has showcased visual tables as robust and generalizable visual representations, laying the groundwork for future research concerning visual reasoning.

\clearpage
\noindent\textbf{Limitations.} 
When training visual table generator, we exclude the VQA annotations to minimize the biases in VQA tasks. However, our generator might still inherit undesired biases from the training datasets and the collected annotations.
Besides, visual tables require computation during generation and downstream tasks. A more efficient way to generate and utilize visual tables can be further explored in future research.

\bibliography{egbib}

%-------------------------------------------------------------------------
% appendix
\clearpage

\appendix

\smallskip
\noindent {\LARGE {{\rm {\bf Appendix}}}}

\bigskip

In the appendix, we provide more details in addition to our main paper: (1) additional details in annotation collection, (2) an introduction to the benchmarks used in experiments, (3) additional implementation details for our MLLMs, (4) additional results using LLMs with smaller size, (5) additional examples of visual reasoning, and (6) additional visualization of our generated visual tables.

%-------------------------------------------------------------------------
\section{Visual Table Annotation Collection}
\begin{figure*}[pt] % [p]
    \centering
    \includegraphics[width=0.98\linewidth]{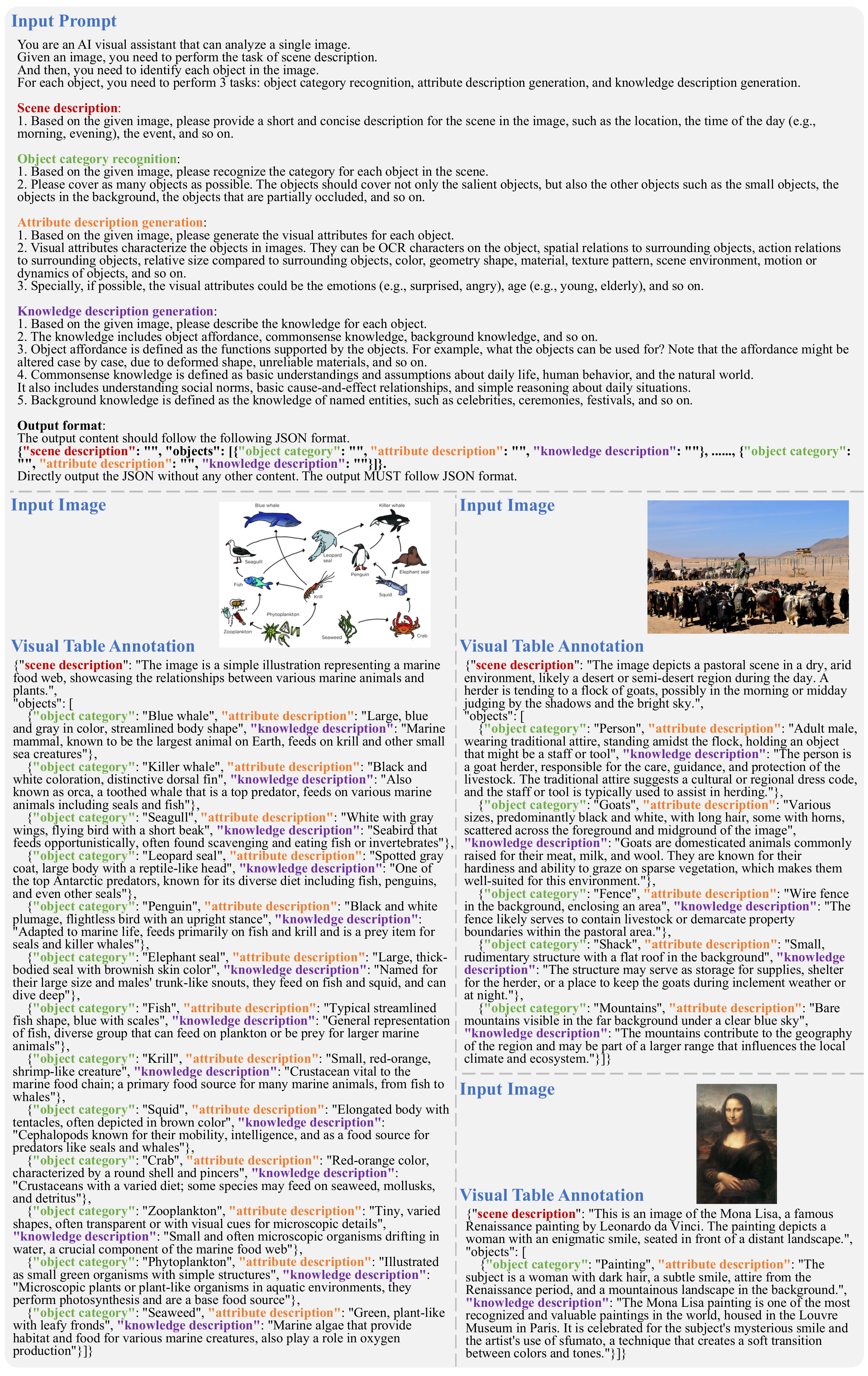}
    \vspace{-8pt}
    \caption{An overview of our designed prompt and the collected visual tables.}
    \label{fig:prompt}
\end{figure*}

To create visual table annotations, instead of collecting costly human annotations, we adopt an alternative solution using powerful foundation models, such as GPT4V \cite{GPT4v}. We create a detailed prompt to ensure the annotations closely align with our requirements, as illustrated in Fig.~\ref{fig:prompt}. 

Without loss of generality, we design visual tables to cover common components in visual environments. 
For instance, the scene description covers time, location, and event, while object descriptions detail the attributes (\eg, color, material, relationships among objects) and world knowledge (\eg, a basic understanding of the physical world and social life). 

Note that this annotation process aims to produce generic representations of visual scenes. No matter human annotators or foundation models, they \textbf{do not} touch any information in downstream tasks, such as the specific ``question'' for an image in VQA task. The resulting benefits on downstream tasks are thus attributed to the design of visual tables, regardless of the way to obtain annotations.

\section{Additional Benchmark Details}

We introduce the benchmarks used in experiments. They are diverse to cover various facets of visual reasoning, including recent benchmarks designed for MLLMs (MM-Vet~\cite{yu2023mm}, LLaVA-Bench~\cite{llava}, MMMU~\cite{yue2023mmmu}, MMBench~\cite{liu2023mmbench}, MMVP~\cite{tong2024mmvp}, POPE~\cite{li2023pope}), and academic VQA benchmarks (VizWiz~\cite{gurari2018vizwiz}, ScienceQA~\cite{lu2022sqa}, GQA~\cite{hudson2019gqa}, VQA-v2~\cite{goyal2017vqav2}, TextVQA~\cite{singh2019textvqa}).

Specifically, MM-Vet assesses six core vision-language capabilities, such as recognition, knowledge, OCR, spatial awareness, language generation, and math. 
LLaVA-Bench evaluates model capabilities on conversation, detailed description, and complex reasoning tasks that usually require world knowledge to answer accurately.
MMMU evaluates models on multi-discipline tasks demanding college-level subject knowledge and deliberate reasoning. 
MMBench evaluates the perception and reasoning capabilities.
MMVP measures visual understanding capabilities by collecting image pairs that CLIP perceives as similar despite their clear visual differences. 
POPE assesses the object hallucination problem. 
VizWiz requires models to answer questions from individuals with visual impairments. 
ScienceQA spans questions from subjects of natural science, language science, and social science. 
GQA and VQA-v2 are traditional VQA benchmarks that evaluate reasoning ability, while GQA focuses more on visual attributes. 
TextVQA evaluates the OCR reasoning capability.

\begin{table*}[h]
\centering
\resizebox{0.70\textwidth}{!}{
\begin{tabular}{l|p{10cm}}
\toprule
\textbf{Benchmark} & \textbf{Response Prompts} \\
\midrule
\midrule
MM-Vet, LLaVA-Bench, MMMU & -- \\
\midrule
POPE, GQA, VQAv2, TextVQA & Answer the question using a single word or phrase. \\
\midrule
MMBench, ScienceQA-IMG, MMVP & Answer with the option's letter from the given choices directly. \\
\midrule
VizWiz & When the provided information is insufficient, respond with `Unanswerable'. Answer the question using a single word or phrase. \\
\bottomrule
\end{tabular}
}
\vspace{-8pt}
\caption{
\textbf{Response prompts} for evaluation benchmarks. The task instruction is concatenated to the input of MLLMs.
}
\label{tab:task_instruction}
% \vspace{-8pt}
% \vspace{-2.0em}
\end{table*}

\begin{table*}[h]
  \centering
  \resizebox{\textwidth}{!}{
  \begin{tabular}{l lcc | c | c | c | c | c | c | c | c | c | c | c }
    \toprule
    Method & LLM & \#PT & \#IT & MM-Vet & LLaVA$^\text{W}$ & MMMU & MMB & MMVP  & POPE & VizWiz & SQA$^\text{I}$ & GQA & VQA$^\text{v2}$ & VQA$^\text{T}$  \\
    \midrule
    \midrule
    \textbf{Representation: E} &  &  &  &  &   &  &  &  &  &  & &  & & \\
    BLIP-2~\cite{li2023blip} & V-13B   & 129M & - & 22.4 & -- & --  & -- & -- & 85.3 & 19.6 & 61.0 & 41.0 & 41.0 & 42.5  \\
    InstructBLIP~\cite{dai2023instructblip} & V-7B   & 129M & 1.2M & 26.2 & -- & -- & 36.0 & -- & -- & 34.5 & 60.5 & 49.2  & -- & 50.1  \\
    InstructBLIP~\cite{dai2023instructblip} & V-13B   & 129M & 1.2M & 25.6 & -- & -- & -- & -- & 78.9 & 33.4 & 63.1 & 49.5 & -- & 50.7  \\
    Shikra~\cite{chen2023shikra} & V-13B   & 600K & 5.5M & -- & --  & -- & 58.8 & -- & -- & -- & -- & -- & 77.4 & -- \\
    IDEFICS-9B~\cite{idefics2023} & L-7B  & 353M & 1M & -- & --  & -- & 48.2 & -- & -- & 35.5 & -- & 38.4 & 50.9 & 25.9  \\
    IDEFICS-80B~\cite{idefics2023} & L-65B & 353M & 1M & -- & -- & -- & 54.5 & -- & -- & 36.0 & -- & 45.2 & 60.0 & 30.9 \\
    Qwen-VL~\cite{bai2023qwen} & Q-7B  & 1.4B  & 50M  & -- & -- & -- & 38.2 & -- & -- & 35.2 & 67.1 & 59.3 & 78.8 & \textbf{63.8} \\
    Qwen-VL-Chat~\cite{bai2023qwen} & Q-7B   & 1.4B   & 50M  & -- & -- & -- & 60.6 & -- & -- & 38.9 & 68.2 & 57.5 & 78.2 & {61.5} \\
    LLaVA-1.5~\cite{liu2023improved} & V-7B  & 558K & 665K & 30.5 & 81.9 & 33.9 & 64.3 & 20.7 & \textbf{85.9} & 50.0 & 66.8 & 62.0 & 78.5 & 58.2 \\
    LLaVA-1.5~\cite{liu2023improved} & V-13B & 558K & 665K & \textbf{35.4} & \textbf{86.7} & \textbf{39.5} & \textbf{67.7} & \textbf{33.3} & \textbf{85.9}  & \textbf{53.6} & \textbf{71.6} & \textbf{63.3} & \textbf{80.0} & 61.3 \\
    \midrule
    \textbf{Representation: T} &   &  &  &  &   &  &  &  &  &  & &  & & \\
    \textbf{Vicuna-VT} & V-7B  & -- & 665K & 28.7 & 82.0 & 33.2 & 61.9 & 19.3 & 81.2 & 52.4 & 67.3 & 55.3 & 73.4 & 51.1  \\
    Vicuna-Cap & V-13B  & -- & 665K & 23.0 & 79.2 & 39.1 & 62.1 & 12.0 & 73.3 & 51.3 & 69.5 & 48.4 & 61.4 & 48.0 \\
    Vicuna-DCap & V-13B  & -- & 665K & 27.1 & 77.6 & 37.4 & 61.4 & 13.3 & \textbf{83.4}  & 51.6 & 69.3 & 51.7 & 68.5 & 49.6 \\
    Vicuna-SG & V-13B  & -- & 665K & 28.1 & 77.0 & 36.5 & 59.3 & 11.3 & 82.3 & 51.0 & 68.9 & 52.0 & 67.8 & 49.3 \\
    \textbf{Vicuna-VT} & V-13B  & -- & 665K & \textbf{30.7} & \textbf{82.5} & \textbf{39.6} & \textbf{62.7} & \textbf{26.7} & 81.9 & \textbf{55.4} & \textbf{70.0} & \textbf{56.1} & \textbf{74.0} & \textbf{53.8}  \\
    \midrule
    \textbf{Representation: E + T} &  &  &  &  &   &  &  &  &  &  & &  & & \\
    \textbf{LLaVA-VT} & V-7B  & 558K & 665K & {31.8} & {85.0} & {34.3} & {68.0} & {24.0} & {86.5} & {50.5} & {67.9} & {63.1} & {79.8} &  59.7 \\
    LLaVA-Cap & V-13B  & 558K & 665K & 36.3 & {88.9} & {40.6} & \textbf{69.4} & 32.0 & 86.4 & 53.8 & {71.9} & {63.5} & {80.3} & 60.7  \\
    LLaVA-DCap & V-13B  & 558K & 665K & {36.7} & 86.4 & 38.6 & 68.9 & 30.7 & 86.9 & 52.6 & 71.2 & 63.4 & {80.3} & 60.7 \\
    LLaVA-SG & V-13B  & 558K & 665K & 36.1 & 86.8 & {40.6} & {69.2} & 30.0  & \textbf{87.3} & \textbf{57.5} & 71.8 & 63.1 & {80.3} & 61.1 \\
    \textbf{LLaVA-VT} & V-13B  & 558K & 665K & \textbf{39.8} & \textbf{89.1} & \textbf{41.9} & \textbf{69.4} & \textbf{36.7} & {87.1} & {57.4} & \textbf{72.6} & \textbf{64.0} & \textbf{80.7} &  \textbf{61.2} \\
    \bottomrule
  \end{tabular}
  }
  \vspace{-8pt}
  \caption{\textbf{Comparison with text-based representations and MLLMs.}  
  \textbf{E}, \textbf{T} and \textbf{E + T} denotes the visual representations as visual embeddings, text-based representations ({Cap}: Short Caption; {Dcap}: Detailed Caption; {SG}: Scene Graph; {VT}: Visual Table) and their concatenation, respectively.
  \#PT/\#IT denotes the number of samples in the stage-one/two training, respectively.
  V-7B/13B: Vicuna-7B/13B~\cite{vicuna2023}; L-7B/13B: LLaMA-7B/13B~\cite{touvron2023llama}; Q-7B/13B: Qwen-7B/13B~\cite{bai2023qwen}. 
  {Bold} values refer to the best results within each group. 
  Visual table \textbf{largely outperforms} previous text-based representations and is the only text representation that can \textbf{consistently} enhance SOTA MLLMs across diverse benchmarks.
  }
  \label{tab:appendix}
  % \vspace{-5pt}
\end{table*}

\section{Additional Implementation Details}

We provide details for response prompts of MLLMs in Tab.~\ref{tab:task_instruction}. Specifically, we follow LLaVA-1.5~\cite{liu2023improved} to design the prompts for different benchmarks, in consideration of the settings of benchmarks. For example, we instruct the MLLM to answer a single word or phrase when the benchmark adopts exact matching evaluation (\ie, VQAv2, GQA, TextVQA, POPE). Similarly, we instruct the MLLM to answer open-ended answers if the benchmark adopts GPT-assisted evaluation (\ie, MM-Vet, LLaVA-Bench, MMMU), to answer the choice letter if the benchmark is evaluated in a multi-choice setting (\ie, MMBench, SQA-IMG, MMVP), and to answer ``Unanswerable'' when the images are unable to provide sufficient information (\ie, VizWiz where many images are blurred).

\section{Additional Experiment Results}

In addition to Tab.~\ref{tab:main} in the main paper, we show more results in Tab.~\ref{tab:appendix} of in appendix. The experiment settings are the same as our main paper, including benchmarks, baselines, and evaluation protocols. The only difference is that we additionally provide results for our MLLMs in 7B size, \ie, Vicuna-VT-7B and LLaVA-VT-7B. 

\smallskip
\noindent \textbf{Vicuna-VT-7B.} Same as the trend in our main paper, visual tables are more effective representations than previous text-based baselines. For example, even if using an LLM with a smaller size, Vicuna-VT-7B can even outperform the baselines that utilize 13B LLMs (\eg, +5.0 on LLaVA-Bench, +2.6 on MMBench, +8.0 on MMVP, +5.6 on VQA-v2 over Vicuna-SG). 
Again, these strong results validate our design of visual tables, providing rich knowledge and precise attributes.

\smallskip
\noindent \textbf{LLaVA-VT-7B.}
Our model LLaVA-VT-7B consistently outperforms the base model LLaVA-1.5-7B across diverse benchmarks. 
Specifically, our model performs better not only on knowledge-intensive benchmarks (\eg, +1.3 on MM-Vet, +3.1 on LLaVA-Bench, +1.1 on SQA), but also on attribute-intensive benchmarks (\eg, +3.4 on MMVP, +1.1 on GQA) and general benchmarks (\eg, +3.7 on MMBench, +1.3 on VQA-v2). 
These promising results again demonstrate that our generated visual tables work as generalizable visual representations, thereby facilitating complex visual reasoning.

\begin{figure*}[t] % [p]
    \centering
    \includegraphics[width=0.99\linewidth]{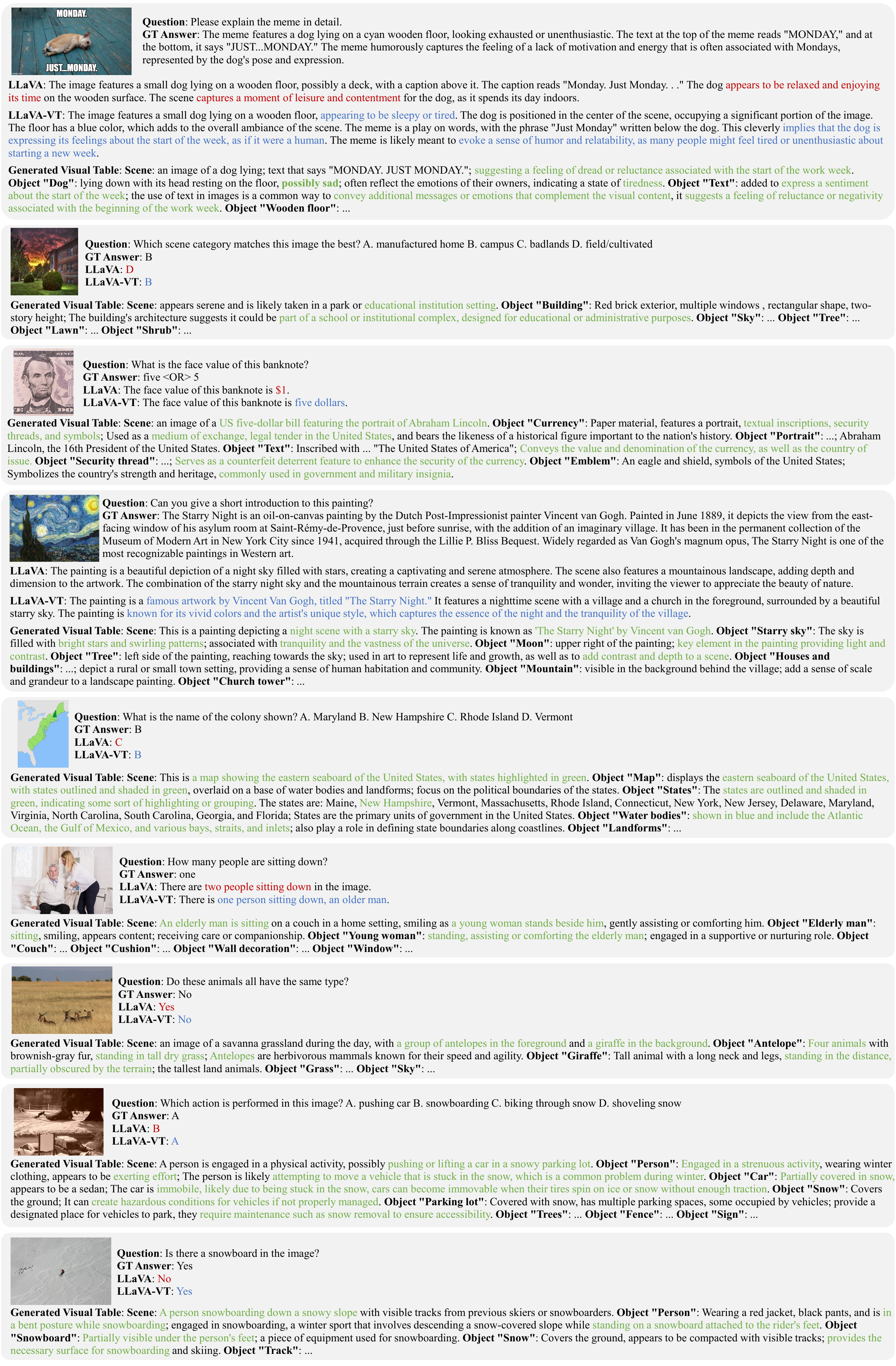}
    \vspace{-8pt}
    \caption{Visualization of visual reasoning examples. For simplicity, we visualize partial visual tables that relate to the questions, with attribute and knowledge separated by ``;''.}
    % \vspace{-10pt}
    \label{fig:case}
\end{figure*}

\section{Additional Visual Reasoning Examples}
With the textual formats, visual tables allow humans to interpret and inspect how they support visual reasoning, as illustrated in Fig.~\ref{fig:case}.
For each image, we visualize the question, the ground-truth answer, the answer from the LLaVA-1.5-13B model, the answer of our MLLM with visual tables, and our generated visual tables. 
For simplicity, we show partial visual tables, \ie, the evidence that supports the visual question answering. We highlight the wrong answers in \textbf{red}, the correct answers in \textbf{blue}, and the supporting evidence from visual tables in \textbf{green}, respectively. 

\textbf{Instance-level knowledge} are provided by visual tables. For instance, the knowledge of ``US five-dollar bill featuring the portrait of Abraham Lincoln'' and ``The Starry Night by Vincent van Gogh'' is given by our visual table for the 3rd and 4th examples, respectively. 
Moreover, our visual table can also provide \textbf{discipline knowledge}, such as the geographic knowledge in the 5th example, ``the eastern seaboard of the United States'' and ``The states are: Maine, New Hampshire, Vermon, ...''.

Besides knowledge, visual tables can \textbf{precisely recognize the object attributes}. For instance, ``Elderly man: sitting; Young woman: standing, assisting or comforting the elderly man'', and ``Antelope: four animals with brownish-gray fur; Giraffe: standing in the distance, partially obscured by the terrain'' are identified in the 6th and 7th examples, respectively, thereby facilitating the correct answer.

Overall, visual tables not only offer interpretability to humans but also thoroughly describe the detailed objects in visual scenes, providing precise attributes and rich knowledge. This comprehensive representation consistently enhances SOTA MLLMs across diverse benchmarks, as demonstrated in our experiments.

\section{Additional Examples of Generated Visual Tables}
In the following tables, we visualize our generated visual tables. As these examples show, our visual tables can thoroughly encode the visual scenes in hierarchical text, by offering concise scene descriptions, accurate object categories, precise object attributes, and rich instance-level object knowledge.

\begin{table*}[t]
\resizebox{\textwidth}{!}{
% \begin{tabular}[]{p{0.5in}p{4.5in}}
\begin{tabular}[]{p{1in}p{4in}}
\toprule
\multicolumn{2}{p{5in}}{
\textbf{Generated Visual Table}}\\
\midrule
\midrule
 & \includegraphics[width=0.95\linewidth]{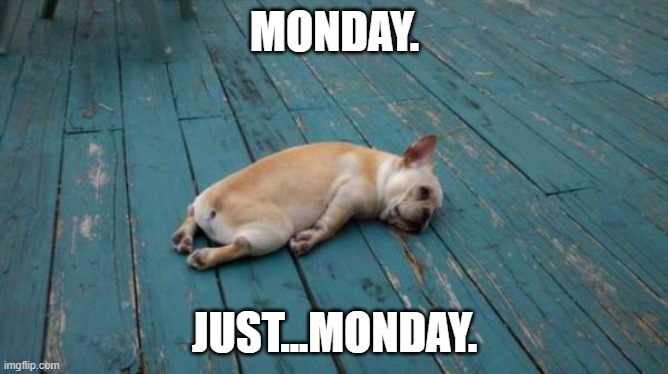} \\
\textbf{User} & \small {Based on the given image, generate the visual table that follows the following JSON format:} \\
& {\small \{\vtred{"scene description"}: "", "objects": [\{\vtobj{"object category"}: "", \vtattr{"attribute description"}: "", \vtknow{"knowledge description"}: ""\}, ......, \{\vtobj{"object category"}: "", \vtattr{"attribute description"}: "", \vtknow{"knowledge description"}: ""\}]\}} \\
\midrule
\textbf{Our Generator} & 
{\small \{\vtred{"scene description"}: "This is an image of a dog lying on a wooden floor, likely inside a house. The image is accompanied by text that says \'MONDAY. JUST MONDAY.\', suggesting a feeling of dread or reluctance associated with the start of the work week.", \newline
"objects": [
\vttab\{\vtobj{"object category"}: "Dog",
\vtattr{"attribute description"}: "The dog appears to be a small breed, possibly a French Bulldog, with a light fawn coat, lying down with its head resting on the floor, looking relaxed or possibly sad.", \vtknow{"knowledge description"}: "Dogs are domesticated animals commonly kept as pets, and they often reflect the emotions of their owners. The posture of the dog suggests it is at rest, possibly indicating a state of relaxation or tiredness."\}, 
\vttab\{\vtobj{"object category"}: "Wooden floor",
\vtattr{"attribute description"}: "The floor has a blue painted surface with visible wood grain and wear, indicating it might be an older or well-used surface.",
\vtknow{"knowledge description"}: "Wooden floors are a common choice for home interiors due to their durability and aesthetic appeal. The wear on the floor suggests it has been well-traveled."\},
\vttab\{\vtobj{"object category"}: "Text",
\vtattr{"attribute description"}: "The text \'MONDAY. JUST MONDAY.\' is superimposed on the image in a sans-serif font, likely digitally added to express a sentiment about the start of the week.",
\vtknow{"knowledge description"}: "The use of text in images is a common way to convey additional messages or emotions that complement the visual content. In this case, it suggests a feeling of reluctance or negativity associated with the beginning of the work week."\}]\}} \\
\bottomrule
\end{tabular}
}
\caption{An example of our generated visual tables.}
\label{table:005}
\end{table*}

\begin{table*}[t]

\resizebox{\textwidth}{!}{
% \begin{tabular}[]{p{0.5in}p{4.5in}}
\begin{tabular}[]{p{1in}p{4in}}
\toprule
\multicolumn{2}{p{5in}}{
\textbf{Generated Visual Table}}\\
\midrule
\midrule
 & \includegraphics[width=0.5\linewidth]{figures/supp/819.jpg} \\
\textbf{User} & \small {Based on the given image, generate the visual table that follows the following JSON format:} \\
& {\small \{\vtred{"scene description"}: "", "objects": [\{\vtobj{"object category"}: "", \vtattr{"attribute description"}: "", \vtknow{"knowledge description"}: ""\}, ......, \{\vtobj{"object category"}: "", \vtattr{"attribute description"}: "", \vtknow{"knowledge description"}: ""\}]\}} \\
\midrule
\textbf{Our Generator} & 
{\small\{\vtred{"scene description"}: "The image depicts an outdoor scene at dusk or dawn with a vibrant pink and orange sky. A building with a distinctive architecture is visible in the background, surrounded by lush greenery and a well-maintained lawn. The scene appears serene and is likely taken in a park or educational institution setting.", "objects": [\vttab\{\vtobj{"object category"}: "Building", \vtattr{"attribute description"}: "Red brick exterior, multiple windows, rectangular shape, two-story height", \vtknow{"knowledge description"}: "The building's architecture suggests it could be part of a school or institutional complex, designed for educational or administrative purposes."\}, \vttab\{\vtobj{"object category"}: "Sky", \vtattr{"attribute description"}: "Pink and orange hues, cloudy, dusk or dawn lighting", \vtknow{"knowledge description"}: "The sky's coloration indicates that the photo was taken during the golden hour, which is shortly after sunrise or before sunset, a time prized for its soft, diffused light."\}, \vttab\{\vtobj{"object category"}: "Tree", \vtattr{"attribute description"}: "Green foliage, medium height, appears healthy", \vtknow{"knowledge description"}: "Trees are commonly found in park settings and are important for providing shade, improving air quality, and enhancing the aesthetic appeal of an area."\}, \vttab\{\vtobj{"object category"}: "Lawn", \vtattr{"attribute description"}: "Green grass, well-maintained, mowed to a medium height", \vtknow{"knowledge description"}: "Lawns are often found in communal outdoor areas and require regular maintenance such as mowing to keep them looking neat."\}, \vttab\{\vtobj{"object category"}: "Shrub", \vtattr{"attribute description"}: "Rounded shape, dense green leaves, medium size compared to the tree", \vtknow{"knowledge description"}: "Shrubs are used in landscaping to create visually appealing boundaries and to add greenery to the environment."\}]\}} \\
\bottomrule
\end{tabular}
}
\caption{An example of our generated visual tables.}
\label{table:819}
\end{table*}

\begin{table*}[t]

\resizebox{\textwidth}{!}{
% \begin{tabular}[]{p{0.5in}p{4.5in}}
\begin{tabular}[]{p{1in}p{4in}}
\toprule
\multicolumn{2}{p{5in}}{
\textbf{Generated Visual Table}}\\
\midrule
\midrule
 & \includegraphics[width=0.5\linewidth]{figures/supp/v1_79.jpg} \\
\textbf{User} & \small {Based on the given image, generate the visual table that follows the following JSON format:} \\
& {\small \{\vtred{"scene description"}: "", "objects": [\{\vtobj{"object category"}: "", \vtattr{"attribute description"}: "", \vtknow{"knowledge description"}: ""\}, ......, \{\vtobj{"object category"}: "", \vtattr{"attribute description"}: "", \vtknow{"knowledge description"}: ""\}]\}} \\
\midrule
\textbf{Our Generator} & 
{\small\{\vtred{"scene description"}: "This is an image of a US five-dollar bill featuring the portrait of Abraham Lincoln.", \newline
"objects": [
\vttab\{\vtobj{"object category"}: "Currency", \vtattr{"attribute description"}: "Paper material, features a portrait, textual inscriptions, security threads, and symbols", \vtknow{"knowledge description"}: "Used as a medium of exchange, legal tender in the United States, and bears the likeness of a historical figure important to the nation's history."\}, \vttab\{\vtobj{"object category"}: "Portrait", \vtattr{"attribute description"}: "Depicts a male with a beard, wearing a black suit and a bow tie", \vtknow{"knowledge description"}: "Represents Abraham Lincoln, the 16th President of the United States, known for leading the country through the Civil War and issuing the Emancipation Proclamation."\}, \vttab\{\vtobj{"object category"}: "Text", \vtattr{"attribute description"}: "Inscribed with 'We Do Not Do' and 'The United States of America'", \vtknow{"knowledge description"}: "Conveys the value and denomination of the currency, as well as the country of issue."\}, \vttab\{\vtobj{"object category"}: "Security thread", \vtattr{"attribute description"}: "Appears as a faint, metallic strand running vertically", \vtknow{"knowledge description"}: "Serves as a counterfeit deterrent feature to enhance the security of the currency."\}, \vttab\{\vtobj{"object category"}: "Emblem", \vtattr{"attribute description"}: "An eagle and shield, symbols of the United States", \vtknow{"knowledge description"}: "Symbolizes the country's strength and heritage, commonly used in government and military insignia."\}]\}}
 \\
\bottomrule
\end{tabular}
}
\caption{An example of our generated visual tables.}
\label{table:v1_79}
\end{table*}

\begin{table*}[t]

\resizebox{\textwidth}{!}{
% \begin{tabular}[]{p{0.5in}p{4.5in}}
\begin{tabular}[]{p{1in}p{4in}}
\toprule
\multicolumn{2}{p{5in}}{
\textbf{Generated Visual Table}}\\
\midrule
\midrule
 & \includegraphics[width=0.45\linewidth]{figures/supp/v1_180.jpg} \\
\textbf{User} & \small {Based on the given image, generate the visual table that follows the following JSON format:} \\
& {\small \{\vtred{"scene description"}: "", "objects": [\{\vtobj{"object category"}: "", \vtattr{"attribute description"}: "", \vtknow{"knowledge description"}: ""\}, ......, \{\vtobj{"object category"}: "", \vtattr{"attribute description"}: "", \vtknow{"knowledge description"}: ""\}]\}} \\
\midrule
\textbf{Our Generator} & 
{\small\{\vtred{"scene description"}: "This is a painting depicting a night scene with a starry sky. The painting is known as 'The Starry Night' by Vincent van Gogh, and it is a representation of a tranquil nighttime landscape with swirling star patterns and a bright moon.", "objects": [\vttab\{\vtobj{"object category"}: "Starry sky", \vtattr{"attribute description"}: "The sky is filled with bright stars and swirling patterns, predominantly blue and yellow in color.", \vtknow{"knowledge description"}: "The starry sky is a representation of the night sky, often associated with tranquility and the vastness of the universe."\}, \vttab\{\vtobj{"object category"}: "Moon", \vtattr{"attribute description"}: "The moon is large and bright, positioned in the upper right of the painting.", \vtknow{"knowledge description"}: "The moon is a celestial body that orbits the Earth and is associated with the night; it is a key element in the painting providing light and contrast."\}, \vttab\{\vtobj{"object category"}: "Tree", \vtattr{"attribute description"}: "A tall, dark green tree stands on the left side of the painting, reaching towards the sky.", \vtknow{"knowledge description"}: "Trees are often used in art to represent life and growth, as well as to add contrast and depth to a scene."\}, \vttab\{\vtobj{"object category"}: "Houses and buildings", \vtattr{"attribute description"}: "Cluster of small, dark buildings with gabled roofs, resembling a village or small town, located below the sky.", \vtknow{"knowledge description"}: "The buildings are likely meant to depict a rural or small town setting, providing a sense of human habitation and community."\}, \vttab\{\vtobj{"object category"}: "Church tower", \vtattr{"attribute description"}: "A tower with a spire, illuminated, stands out against the dark sky, located in the center of the village.", \vtknow{"knowledge description"}: "Church towers are often focal points in a village, serving as landmarks and symbols of community and spirituality."\}, \vttab\{\vtobj{"object category"}: "Mountain", \vtattr{"attribute description"}: "A large, dark mountain or hill is visible in the background behind the village.", \vtknow{"knowledge description"}: "Mountains are natural landforms that often add a sense of scale and grandeur to a landscape painting."\}]\}}
 \\
\bottomrule
\end{tabular}
}
\caption{An example of our generated visual tables.}
\label{table:v1_180}
\end{table*}

\begin{table*}[t]

\resizebox{\textwidth}{!}{
% \begin{tabular}[]{p{0.5in}p{4.5in}}
\begin{tabular}[]{p{1in}p{4in}}
\toprule
\multicolumn{2}{p{5in}}{
\textbf{Generated Visual Table}}\\
\midrule
\midrule
 & \includegraphics[width=0.3\linewidth]{figures/supp/11_image.png} \\
\textbf{User} & \small {Based on the given image, generate the visual table that follows the following JSON format:} \\
& {\small \{\vtred{"scene description"}: "", "objects": [\{\vtobj{"object category"}: "", \vtattr{"attribute description"}: "", \vtknow{"knowledge description"}: ""\}, ......, \{\vtobj{"object category"}: "", \vtattr{"attribute description"}: "", \vtknow{"knowledge description"}: ""\}]\}} \\
\midrule
\textbf{Our Generator} & 
{\small\{\vtred{"scene description"}: "This is a map showing the eastern seaboard of the United States, with states highlighted in green. The map appears to be a political map rather than a physical or geographical map.", \newline
"objects": [
\vttab\{\vtobj{"object category"}: "Map", \vtattr{"attribute description"}: "The map displays the eastern seaboard of the United States, with states outlined and shaded in green. The map includes state names and is overlaid on a base of water bodies and landforms.", \vtknow{"knowledge description"}: "Maps are used for navigation, geography education, and reference. This particular map seems to focus on the political boundaries of the states rather than physical features, which is common for political or election-related purposes."\}, \vttab\{\vtobj{"object category"}: "States", \vtattr{"attribute description"}: "The states are outlined and shaded in green, indicating some sort of highlighting or grouping. The states are: Maine, New Hampshire, Vermont, Massachusetts, Rhode Island, Connecticut, New York, New Jersey, Delaware, Maryland, Virginia, North Carolina, South Carolina, Georgia, and Florida.", \vtknow{"knowledge description"}: "States are the primary units of government in the United States, each with its own set of laws, governors, and legislative bodies. The highlighting of states could be for electoral purposes, regional planning, or educational material."\}, \vttab\{\vtobj{"object category"}: "Water bodies", \vtattr{"attribute description"}: "The water bodies are shown in blue and include the Atlantic Ocean, the Gulf of Mexico, and various bays, straits, and inlets.", \vtknow{"knowledge description"}: "Water bodies are essential for transportation, recreation, and are a vital part of the ecosystem. They also play a role in defining state boundaries along coastlines."\}, \vttab\{\vtobj{"object category"}: "Landforms", \vtattr{"attribute description"}: "The landforms are shown in various shades, indicating different elevations or terrain types.", \vtknow{"knowledge description"}: "Landforms are natural features of the Earth's surface and are important for understanding geography, climate, and ecosystems."\}]\}}
 \\
\bottomrule
\end{tabular}
}
\caption{An example of our generated visual tables.}
\label{table:11_image}
\end{table*}

\begin{table*}[t]

\resizebox{\textwidth}{!}{
% \begin{tabular}[]{p{0.5in}p{4.5in}}
\begin{tabular}[]{p{1in}p{4in}}
\toprule
\multicolumn{2}{p{5in}}{
\textbf{Generated Visual Table}}\\
\midrule
\midrule
 & \includegraphics[width=0.50\linewidth]{figures/supp/v1_108.jpg} \\
\textbf{User} & \small {Based on the given image, generate the visual table that follows the following JSON format:} \\
& {\small \{\vtred{"scene description"}: "", "objects": [\{\vtobj{"object category"}: "", \vtattr{"attribute description"}: "", \vtknow{"knowledge description"}: ""\}, ......, \{\vtobj{"object category"}: "", \vtattr{"attribute description"}: "", \vtknow{"knowledge description"}: ""\}]\}} \\
\midrule
\textbf{Our Generator} & 
{\small\{\vtred{"scene description"}: "An elderly man is sitting on a couch in a home setting, smiling as a young woman stands beside him, gently assisting or comforting him. The environment suggests a domestic, caring interaction, possibly in a living room during the day.", "objects": [
\vttab\{\vtobj{"object category"}: "Elderly man", \vtattr{"attribute description"}: "sitting, smiling, wearing a cream-colored sweater, white hair, appears content", \vtknow{"knowledge description"}: "The elderly man is likely a family member or friend, receiving care or companionship. His attire and the home setting suggest a comfortable, familiar environment."\}, \vttab\{\vtobj{"object category"}: "Young woman", \vtattr{"attribute description"}: "standing, blonde hair, wearing a white shirt and dark pants, appears to be assisting or comforting the elderly man", \vtknow{"knowledge description"}: "The young woman is likely a family member, caregiver, or friend, engaged in a supportive or nurturing role. Her attire is casual, indicating an informal setting."\}, \vttab\{\vtobj{"object category"}: "Couch", \vtattr{"attribute description"}: "striped cushion, part of the furniture in the room", \vtknow{"knowledge description"}: "The couch is a piece of furniture designed for seating, commonly found in living rooms and used for relaxation or socializing."\}, \vttab\{\vtobj{"object category"}: "Cushion", \vtattr{"attribute description"}: "striped pattern, placed on the couch", \vtknow{"knowledge description"}: "Cushions are used for added comfort and decoration on furniture such as couches and chairs."\}, \vttab\{\vtobj{"object category"}: "Wall decoration", \vtattr{"attribute description"}: "white, appears to be a decorative item on the wall", \vtknow{"knowledge description"}: "Wall decorations are used to enhance the aesthetic appeal of a room and often reflect personal tastes."\}, \vttab\{\vtobj{"object category"}: "Window", \vtattr{"attribute description"}: "partially visible, covered with a sheer curtain", \vtknow{"knowledge description"}: "Windows allow natural light to enter the room and can be covered with curtains for privacy or to control the amount of light."\}]\}}
 \\
\bottomrule
\end{tabular}
}
\caption{An example of our generated visual tables.}
\label{table:v1_108}
\end{table*}

\begin{table*}[t]

\resizebox{\textwidth}{!}{
% \begin{tabular}[]{p{0.5in}p{4.5in}}
\begin{tabular}[]{p{1in}p{4in}}
\toprule
\multicolumn{2}{p{5in}}{
\textbf{Generated Visual Table}}\\
\midrule
\midrule
 & \includegraphics[width=0.9\linewidth]{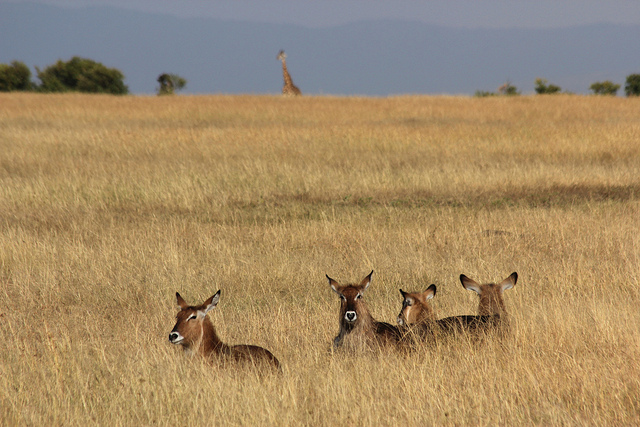} \\
\textbf{User} & \small {Based on the given image, generate the visual table that follows the following JSON format:} \\
& {\small \{\vtred{"scene description"}: "", "objects": [\{\vtobj{"object category"}: "", \vtattr{"attribute description"}: "", \vtknow{"knowledge description"}: ""\}, ......, \{\vtobj{"object category"}: "", \vtattr{"attribute description"}: "", \vtknow{"knowledge description"}: ""\}]\}} \\
\midrule
\textbf{Our Generator} & 
{\small\{\vtred{"scene description"}: "This is an image of a savanna grassland during the day, with a group of antelopes in the foreground and a giraffe in the background.", \newline
"objects": [
\vttab\{\vtobj{"object category"}: "Antelope", \vtattr{"attribute description"}: "Four animals with brownish-gray fur, visible horns, ears pointed upwards, standing in tall dry grass.", \vtknow{"knowledge description"}: "Antelopes are herbivorous mammals known for their speed and agility. They live in herds and are commonly found in savanna ecosystems where they graze on grasses and leaves."\}, \vttab\{\vtobj{"object category"}: "Giraffe", \vtattr{"attribute description"}: "Tall animal with a long neck and legs, spotted pattern, standing in the distance, partially obscured by the terrain.", \vtknow{"knowledge description"}: "Giraffes are the tallest land animals, adapted to browsing foliage higher than other herbivores can reach, usually found in African savanna environments."\}, \vttab\{\vtobj{"object category"}: "Grass", \vtattr{"attribute description"}: "Tall, dry, yellowish grass covering the ground.", \vtknow{"knowledge description"}: "Grass in savannas is adapted to dry conditions and periodic fires, playing a crucial role in the ecosystem as a food source for herbivores."\}, \vttab\{\vtobj{"object category"}: "Sky", \vtattr{"attribute description"}: "Clear blue sky with no visible clouds.", \vtknow{"knowledge description"}: "The clear sky suggests fair weather, which is common in savanna climates that have distinct wet and dry seasons."\}]\}}
 \\
\bottomrule
\end{tabular}
}
\caption{An example of our generated visual tables.}
\label{table:n469525}
\end{table*}

\begin{table*}[t]

% \resizebox{\textwidth}{!}{
\resizebox{\textwidth}{!}{
% \begin{tabular}[]{p{0.5in}p{4.5in}}
\begin{tabular}[]{p{1in}p{4in}}
\toprule
\multicolumn{2}{p{5in}}{
\textbf{Generated Visual Table}}\\
\midrule
\midrule
 & \includegraphics[width=0.40\linewidth]{figures/supp/594.jpg} \\
\textbf{User} & \small {Based on the given image, generate the visual table that follows the following JSON format:} \textbf{......}\\
\midrule
\textbf{Our Generator} & 
{\small\{\vtred{"scene description"}: "A person is engaged in a physical activity, possibly pushing or lifting a car in a snowy parking lot during the daytime.", "objects": [\vttab\{\vtobj{"object category"}: "Person", \vtattr{"attribute description"}: "Engaged in a strenuous activity, wearing winter clothing, appears to be exerting effort", \vtknow{"knowledge description"}: "The person is likely attempting to move a vehicle that is stuck in the snow, which is a common problem during winter. This requires physical strength and can be a challenging task."\}, \vttab\{\vtobj{"object category"}: "Car", \vtattr{"attribute description"}: "Partially covered in snow, stationary, appears to be a sedan", \vtknow{"knowledge description"}: "The car is immobile, likely due to being stuck in the snow. Cars can become immovable when their tires spin on ice or snow without enough traction."\}, \vttab\{\vtobj{"object category"}: "Snow", \vtattr{"attribute description"}: "Covers the ground, white, appears to be compacted in areas", \vtknow{"knowledge description"}: "Snow is a form of precipitation that occurs when temperatures are low enough to allow water vapor to solidify. It can create hazardous conditions for vehicles if not properly managed."\}, \vttab\{\vtobj{"object category"}: "Trees", \vtattr{"attribute description"}: "Tall, leafless, possibly dormant due to winter season", \vtknow{"knowledge description"}: "Trees lose their leaves in the winter to conserve water and energy. They are often found in parking lots to provide aesthetic value and shade during warmer months."\}, \vttab\{\vtobj{"object category"}: "Fence", \vtattr{"attribute description"}: "White, wooden, appears to be a boundary marker", \vtknow{"knowledge description"}: "Fences are used to mark property boundaries, provide security, and sometimes to prevent vehicles from entering certain areas."\}, \vttab\{\vtobj{"object category"}: "Parking lot", \vtattr{"attribute description"}: "Covered with snow, has multiple parking spaces, some occupied by vehicles", \vtknow{"knowledge description"}: "Parking lots are common in urban and suburban areas to provide a designated place for vehicles to park. They require maintenance such as snow removal to ensure accessibility."\}, \vttab\{\vtobj{"object category"}: "Sign", \vtattr{"attribute description"}: "White, appears to be a regulatory or informational sign", \vtknow{"knowledge description"}: "Signs in parking lots provide rules, directions, or information to drivers and pedestrians to maintain order and safety."\}]\}}
 \\
\bottomrule
\end{tabular}
}
\caption{An example of our generated visual tables.}
\label{table:594}
\end{table*}

\begin{table*}[t]

\resizebox{\textwidth}{!}{
% \begin{tabular}[]{p{0.5in}p{4.5in}}
\begin{tabular}[]{p{1in}p{4in}}
\toprule
\multicolumn{2}{p{5in}}{
\textbf{Generated Visual Table}}\\
\midrule
\midrule
 & \includegraphics[width=0.9\linewidth]{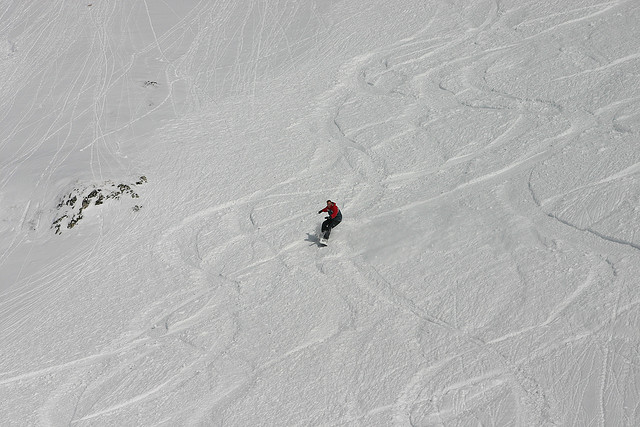} \\
\textbf{User} & \small {Based on the given image, generate the visual table that follows the following JSON format:} \\
& {\small \{\vtred{"scene description"}: "", "objects": [\{\vtobj{"object category"}: "", \vtattr{"attribute description"}: "", \vtknow{"knowledge description"}: ""\}, ......, \{\vtobj{"object category"}: "", \vtattr{"attribute description"}: "", \vtknow{"knowledge description"}: ""\}]\}} \\
\midrule
\textbf{Our Generator} & 
{\small\{\vtred{"scene description"}: "A person snowboarding down a snowy slope with visible tracks from previous skiers or snowboarders.", \newline
"objects": [
\vttab\{\vtobj{"object category"}: "Person", \vtattr{"attribute description"}: "Wearing a red jacket, black pants, and is in a bent posture while snowboarding", \vtknow{"knowledge description"}: "The person is engaged in snowboarding, a winter sport that involves descending a snow-covered slope while standing on a snowboard attached to the rider's feet."\}, \vttab\{\vtobj{"object category"}: "Snowboard", \vtattr{"attribute description"}: "Partially visible under the person's feet", \vtknow{"knowledge description"}: "A snowboard is a piece of equipment used for snowboarding. It is designed to glide over snow and allows the rider to maneuver through turns and tricks."\}, \vttab\{\vtobj{"object category"}: "Snow", \vtattr{"attribute description"}: "Covers the ground, appears to be compacted with visible tracks", \vtknow{"knowledge description"}: "Snow is a form of precipitation that occurs when temperatures are low enough to allow water vapor to solidify. It provides the necessary surface for snowboarding and skiing."\}, \vttab\{\vtobj{"object category"}: "Tracks", \vtattr{"attribute description"}: "Curved lines in the snow made by previous skiers or snowboarders", \vtknow{"knowledge description"}: "Tracks in the snow are left by the passage of skiers or snowboarders. They indicate the popularity of the slope and the skill of the individuals as tracks can vary in depth and width based on the activity and conditions."\}]\}}
 \\
\bottomrule
\end{tabular}
}
\caption{An example of our generated visual tables.}
\label{table:000000310196}
\end{table*}

\begin{table*}[t]

\resizebox{\textwidth}{!}{
% \begin{tabular}[]{p{0.5in}p{4.5in}}
\begin{tabular}[]{p{1in}p{4in}}
\toprule
\multicolumn{2}{p{5in}}{
\textbf{Generated Visual Table}}\\
\midrule
\midrule
 & \includegraphics[width=0.7\linewidth]{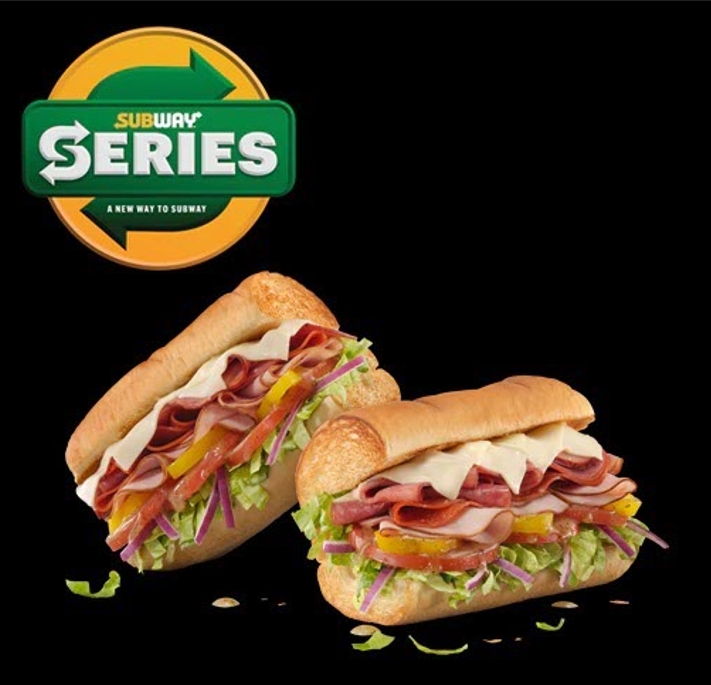} \\
\textbf{User} & \small {Based on the given image, generate the visual table that follows the following JSON format:} \\
& {\small \{\vtred{"scene description"}: "", "objects": [\{\vtobj{"object category"}: "", \vtattr{"attribute description"}: "", \vtknow{"knowledge description"}: ""\}, ......, \{\vtobj{"object category"}: "", \vtattr{"attribute description"}: "", \vtknow{"knowledge description"}: ""\}]\}} \\
\midrule
\textbf{Our Generator} & 
{\small\{\vtred{"scene description"}: "This is an image of a food item, specifically a sandwich, against a black background with a logo in the top left corner.", \newline
"objects": [
\vttab\{\vtobj{"object category"}: "Sandwich", \vtattr{"attribute description"}: "The sandwich appears to have a variety of fillings including meats, cheese, lettuce, and tomatoes, placed between two slices of bread with visible seeds on the bread crusts.", \vtknow{"knowledge description"}: "A sandwich is a common food item typically consisting of vegetables, sliced cheese or meat, placed on or between slices of bread. It is a popular choice for meals and can be found in many cultures worldwide. Sandwiches are known for their convenience as they are easy to prepare and can be customized with a variety of ingredients to cater to different tastes."\}, \vttab\{\vtobj{"object category"}: "Logo", \vtattr{"attribute description"}: "The logo is a circular emblem with a green outline and the text 'Subway Series', along with a stylized 'S' in the center. It is placed against a black background.", \vtknow{"knowledge description"}: "The logo is likely associated with the Subway fast food restaurant chain, which is known for its submarine sandwiches. The term 'Series' might suggest a new promotion or a specific menu line. The 'S' could be a stylized representation of the brand's initial."\}]\}}
 \\
\bottomrule
\end{tabular}
}
\caption{An example of our generated visual tables.}
\label{table:023}
\end{table*}

\begin{table*}[t]

\resizebox{\textwidth}{!}{
% \begin{tabular}[]{p{0.5in}p{4.5in}}
\begin{tabular}[]{p{1in}p{4in}}
\toprule
\multicolumn{2}{p{5in}}{
\textbf{Generated Visual Table}}\\
\midrule
\midrule
 & \includegraphics[width=0.4\linewidth]{figures/supp/v1_41.png} \\
\textbf{User} & \small {Based on the given image, generate the visual table that follows the following JSON format:} \textbf{......}\\
\midrule
\textbf{Our Generator} & 
{\small\{\vtred{"scene description"}: "This image is a collage of nine squares, each containing a different strategy or tip related to memory and learning, likely intended for educational purposes.", "objects": [\vttab\{\vtobj{"object category"}: "Square 1", \vtattr{"attribute description"}: "The image shows a clip art of a pen and paper with the text 'Make lists'. The items are in black and white, and the text is in a bold, sans-serif font.", \vtknow{"knowledge description"}: "The pen and paper represent writing tools commonly used for note-taking and organizing information. Making lists is a memory strategy that aids in information retention and retrieval."\}, \vttab\{\vtobj{"object category"}: "Square 2", \vtattr{"attribute description"}: "This square features a light bulb clip art with the text 'Make associations'. The light bulb is stylized with a filament inside, and the text is in a bold, serif font.", \vtknow{"knowledge description"}: "The light bulb is a symbol for an idea or insight. Making associations involves connecting new information with existing knowledge to enhance memory."\}, \vttab\{\vtobj{"object category"}: "Square 3", \vtattr{"attribute description"}: "This square includes a clip art of a calendar with the text 'Review regularly'. The calendar is simple with just the days of the week visible, and the text is in a bold, serif font.", \vtknow{"knowledge description"}: "A calendar is a tool for keeping track of dates and can be used for scheduling regular review sessions to reinforce learning."\}, \vttab\{\vtobj{"object category"}: "Square 4", \vtattr{"attribute description"}: "The image shows a clip art of a person with a checklist, labeled 'Visualize'. The person is depicted with a thought bubble, and the text is in a bold, serif font.", \vtknow{"knowledge description"}: "Visualization involves creating mental images to aid in memory and can be used to simulate real-life scenarios for better understanding and retention."\}, 
\vttab \textbf{......,}
\vttab\{\vtobj{"object category"}: "Square 9", \vtattr{"attribute description"}: "This square features a clip art of a person with a music note, labeled 'Make a song out of it'. The person is playing a musical instrument, and the text is in a bold, serif font.", \vtknow{"knowledge description"}: "Turning information into a song can be a creative way to remember information through melody and rhythm."\}]\}}
 \\
\bottomrule
\end{tabular}
}
\caption{An example of our generated visual tables.}
\label{table:v1_41}
\end{table*}

\begin{table*}[t]

\resizebox{\textwidth}{!}{
% \begin{tabular}[]{p{0.5in}p{4.5in}}
\begin{tabular}[]{p{1in}p{4in}}
\toprule
\multicolumn{2}{p{5in}}{
\textbf{Generated Visual Table}}\\
\midrule
\midrule
 & \includegraphics[width=0.6\linewidth]{figures/supp/v1_99.png} \\
\textbf{User} & \small {Based on the given image, generate the visual table that follows the following JSON format:} ...\\
\midrule
\textbf{Our Generator} & 
{\small\{\vtred{"scene description"}: "This is an illustration of a marine ecosystem with various sea creatures and plants, accompanied by arrows indicating feeding relationships and labels for each organism.", "objects": [\vttab\{\vtobj{"object category"}: "Blue whale", \vtattr{"attribute description"}: "Large, blue, mammal, labeled 'Killer whale'", \vtknow{"knowledge description"}: "The blue whale is the largest mammal in the world, known for its size and for eating other large marine animals."\}, \vttab\{\vtobj{"object category"}: "Killer whale", \vtattr{"attribute description"}: "Large, black and white, mammal, labeled 'Blue whale'", \vtknow{"knowledge description"}: "Killer whales are known for their powerful hunting abilities and are often found in groups."\}, \vttab\{\vtobj{"object category"}: "Leopard seal", \vtattr{"attribute description"}: "Large, seal, spotted pattern, labeled 'Krill'", \vtknow{"knowledge description"}: "Seals are semi-aquatic marine mammals that feed on a variety of prey, including krill."\}, \vttab\{\vtobj{"object category"}: "Krill", \vtattr{"attribute description"}: "Small, red, crustacean, labeled 'Leopard seal'", \vtknow{"knowledge description"}: "Krill are small, shrimp-like creatures that are a staple in the diet of many marine animals."\}, \vttab\{\vtobj{"object category"}: "Penguin", \vtattr{"attribute description"}: "Medium-sized, black and white, bird, labeled 'Penguin'", \vtknow{"knowledge description"}: "Penguins are flightless birds that are adapted for swimming in the ocean."\}, \vttab\{\vtobj{"object category"}: "Seaweed", \vtattr{"attribute description"}: "Green, algae, labeled 'Seaweed'", \vtknow{"knowledge description"}: "Seaweed refers to various types of marine plants and algae that live in the ocean."\}, 
\vttab\{\vtobj{"object category"}: "Phytoplankton", \vtattr{"attribute description"}: "Microscopic, green, labeled 'Phytoplankton'", \vtknow{"knowledge description"}: "Phytoplankton are tiny, photosynthetic organisms that are the base of the marine food web."\}, 
\vttab \textbf{......,}
\vttab\{\vtobj{"object category"}: "Sea star", \vtattr{"attribute description"}: "Medium-sized, star-shaped, labeled 'Sea star'", \vtknow{"knowledge description"}: "Sea stars are echinoderms with a distinctive star shape and are found in a variety of marine environments."\}]\}}
 \\
\bottomrule
\end{tabular}
}
\caption{An example of our generated visual tables.}
\label{table:v1_99}
\end{table*}

\begin{table*}[t]

\resizebox{\textwidth}{!}{
% \begin{tabular}[]{p{0.5in}p{4.5in}}
\begin{tabular}[]{p{1in}p{4in}}
\toprule
\multicolumn{2}{p{5in}}{
\textbf{Generated Visual Table}}\\
\midrule
\midrule
 & \includegraphics[width=0.6\linewidth]{figures/supp/v1_83.png} \\
\textbf{User} & \small {Based on the given image, generate the visual table that follows the following JSON format:} \\
& {\small \{\vtred{"scene description"}: "", "objects": [\{\vtobj{"object category"}: "", \vtattr{"attribute description"}: "", \vtknow{"knowledge description"}: ""\}, ......, \{\vtobj{"object category"}: "", \vtattr{"attribute description"}: "", \vtknow{"knowledge description"}: ""\}]\}} \\
\midrule
\textbf{Our Generator} & 
{\small\{\vtred{"scene description"}: "This is an X-ray image of a person's chest, highlighting the internal structures of the ribcage and the lungs.", \newline
"objects": [
\vttab\{\vtobj{"object category"}: "Ribcage", \vtattr{"attribute description"}: "Bony structure, visible as the curved skeletal framework encompassing the chest cavity, white color due to the X-ray image", \vtknow{"knowledge description"}: "The ribcage provides protection for the heart, lungs, and other internal organs. It is also involved in the process of breathing, as the ribs play a role in expanding the chest to inhale air."\}, \vttab\{\vtobj{"object category"}: "Lungs", \vtattr{"attribute description"}: "Pair of spongy, air-filled organs, visible as light areas within the darker background, indicating the presence of air spaces", \vtknow{"knowledge description"}: "The lungs are responsible for gas exchange, taking in oxygen and expelling carbon dioxide. They are essential for breathing and are a key part of the respiratory system."\}, \vttab\{\vtobj{"object category"}: "X-ray label", \vtattr{"attribute description"}: "Characters 'J.8H', likely indicating a code or identifier for the X-ray image, placed in the top right corner", \vtknow{"knowledge description"}: "X-ray labels are used for patient identification and to help organize and retrieve medical images in a radiology department's database."\}]\}}
 \\
\bottomrule
\end{tabular}
}
\caption{An example of our generated visual tables.}
\label{table:v1_83}
\end{table*}

\begin{table*}[t]

\resizebox{\textwidth}{!}{
% \begin{tabular}[]{p{0.5in}p{4.5in}}
\begin{tabular}[]{p{1in}p{4in}}
\toprule
\multicolumn{2}{p{5in}}{
\textbf{Generated Visual Table}}\\
\midrule
\midrule
 & \includegraphics[width=0.9\linewidth]{figures/supp/validation_Biology_3.png} \\
\textbf{User} & \small {Based on the given image, generate the visual table that follows the following JSON format:} \\
& {\small \{\vtred{"scene description"}: "", "objects": [\{\vtobj{"object category"}: "", \vtattr{"attribute description"}: "", \vtknow{"knowledge description"}: ""\}, ......, \{\vtobj{"object category"}: "", \vtattr{"attribute description"}: "", \vtknow{"knowledge description"}: ""\}]\}} \\
\midrule
\textbf{Our Generator} & 
{\small\{\vtred{"scene description"}: "This is an illustration of extracellular fluid, specifically showing a section of connective tissue with labeled components.", "objects": [\vttab\{\vtobj{"object category"}: "Extracellular fluid", \vtattr{"attribute description"}: "Surrounding the connective tissue cells, transparent, with label 'Extracellular fluid'", \vtknow{"knowledge description"}: "The extracellular fluid is the fluid that is outside of the cells and bathes the cells, playing a crucial role in maintaining the balance of bodily fluids and transporting nutrients and waste products."\}, \vttab\{\vtobj{"object category"}: "Connective tissue cells", \vtattr{"attribute description"}: "Yellow with protrusions, labeled 'Connective tissue cells'", \vtknow{"knowledge description"}: "Connective tissue cells are a diverse group of cells that provide structural support and are involved in the repair and regeneration of damaged tissues."\}, \vttab\{\vtobj{"object category"}: "Glucose", \vtattr{"attribute description"}: "Small yellow squares with arrows indicating movement, labeled 'Glucose'", \vtknow{"knowledge description"}: "Glucose is a type of sugar that serves as a primary source of energy for cells and is transported through the bloodstream."\}, \vttab\{\vtobj{"object category"}: "Cytoplasm", \vtattr{"attribute description"}: "Visible within the connective tissue cells, labeled 'Cytoplasm'", \vtknow{"knowledge description"}: "The cytoplasm is the gelatinous interior of a cell, where metabolic processes occur, and contains various organelles and enzymes necessary for cell function."\}, \vttab\{\vtobj{"object category"}: "Labels", \vtattr{"attribute description"}: "Text labels in English, 'Extracellular fluid', 'Connective tissue cells', 'Glucose', 'Cytoplasm'", \vtknow{"knowledge description"}: "Labels are used in scientific illustrations to identify and explain the functions and components of the depicted structures."\}]\}}
 \\
\bottomrule
\end{tabular}
}
\caption{An example of our generated visual tables.}
\label{table:validation_Biology_3}
\end{table*}

\begin{table*}[t]

\resizebox{\textwidth}{!}{
% \begin{tabular}[]{p{0.5in}p{4.5in}}
\begin{tabular}[]{p{1in}p{4in}}
\toprule
\multicolumn{2}{p{5in}}{
\textbf{Generated Visual Table}}\\
\midrule
\midrule
 & \includegraphics[width=0.6\linewidth]{figures/supp/validation_Biology_19.png} \\
\textbf{User} & \small {Based on the given image, generate the visual table that follows the following JSON format:} \textbf{......}\\
\midrule
\textbf{Our Generator} & 
{\small\{\vtred{"scene description"}: "This is an illustration of the process of bacterial transformation, showing a bacterial cell with foreign DNA being taken up, followed by the integration of the foreign DNA into the bacterial chromosome, and the expression of the new traits conferred by the acquired DNA.", "objects": [\vttab\{\vtobj{"object category"}: "Bacterial cell", \vtattr{"attribute description"}: "Ellipsoidal shape, outer membrane, cytoplasm, plasmid with replication bubbles", \vtknow{"knowledge description"}: "Bacterial cells are basic units of bacteria, which are single-celled microorganisms. They have a cell wall, genetic material in form of DNA, and they replicate through a process that includes replication and transcription."\}, \vttab\{\vtobj{"object category"}: "Plasmid", \vtattr{"attribute description"}: "Small, circular DNA molecule within the bacterial cell, visible replication sites indicated by arrows", \vtknow{"knowledge description"}: "Plasmids are extra-chromosomal, self-replicating genetic elements that are present in many bacteria. They can carry genes that confer antibiotic resistance or other beneficial traits for the bacteria."\}, \vttab\{\vtobj{"object category"}: "Foreign DNA", \vtattr{"attribute description"}: "Linear molecule, outside the bacterial cell, being taken up by the cell", \vtknow{"knowledge description"}: "Foreign DNA can be introduced into bacterial cells through various means, such as transformation. Once inside, the DNA may integrate into the bacterial chromosome or exist as a separate plasmid."\}, \vttab\{\vtobj{"object category"}: "Bacterial transformation", \vtattr{"attribute description"}: "Uptake of foreign DNA by the bacterial cell, process indicated by arrows and a taken-up fragment of DNA", \vtknow{"knowledge description"}: "Bacterial transformation is the process by which bacteria take up external genetic material, which may provide them with new traits or genes that enhance their survival."\}, \vttab\{\vtobj{"object category"}: "Chromosome", \vtattr{"attribute description"}: "Circular structure, incorporating the foreign DNA, indicating replication", \vtknow{"knowledge description"}: "The bacterial chromosome is the main genetic material of the cell, where the majority of genes are located. It replicates to ensure that each daughter cell receives a copy of the genetic material."\}, \vttab\{\vtobj{"object category"}: "New traits", \vtattr{"attribute description"}: "Traits conferred by the acquired DNA, indicated by the new morphology and labeled 'New Bacterial DNA'", \vtknow{"knowledge description"}: "The acquisition of new genetic material through transformation can result in the expression of novel traits in the bacteria, such as antibiotic resistance or the ability to utilize new types of substrates."\}]\}}
 \\
\bottomrule
\end{tabular}
}
\caption{An example of our generated visual tables.}
\label{table:validation_Biology_19}
\end{table*}

\begin{table*}[t]

\resizebox{\textwidth}{!}{
% \begin{tabular}[]{p{0.5in}p{4.5in}}
\begin{tabular}[]{p{1in}p{4in}}
\toprule
\multicolumn{2}{p{5in}}{
\textbf{Generated Visual Table}}\\
\midrule
\midrule
 & \includegraphics[width=0.4\linewidth]{figures/supp/v1_185.jpg} \\
\textbf{User} & \small {Based on the given image, generate the visual table that follows the following JSON format:} \textbf{......}\\
\midrule
\textbf{Our Generator} & 
{\small\{\vtred{"scene description"}: "This is an East Asian-style painting depicting a bustling scene at a bridge over a moat with people and boats. The setting appears to be in a historical or mythological context, with traditional architecture visible in the background.", "objects": [\vttab\{\vtobj{"object category"}: "Bridge", \vtattr{"attribute description"}: "Large, arched, wooden structure with people on it", \vtknow{"knowledge description"}: "Bridges are structures built to span physical obstacles without closing the way underneath such as a body of water, valley, or road, for the purpose of providing passage over the obstacle."\}, \vttab\{\vtobj{"object category"}: "Moat", \vtattr{"attribute description"}: "Wide, water-filled ditch surrounding the area", \vtknow{"knowledge description"}: "Moats are often used as protective barriers around castles, cities, or other structures, and can also be part of a landscape design."\}, \vttab\{\vtobj{"object category"}: "People", \vtattr{"attribute description"}: "Various individuals depicted in traditional clothing, engaged in different activities such as walking and boating", \vtknow{"knowledge description"}: "The people are likely depicted in a style that reflects the customs and societal roles of the time period or cultural setting in which the scene is meant to take place."\}, \vttab\{\vtobj{"object category"}: "Boats", \vtattr{"attribute description"}: "Small, traditional-looking boats on the water", \vtknow{"knowledge description"}: "Boats are used for transportation across water and can vary greatly in size and design based on their intended use and cultural context."\}, 
\vttab \textbf{......,}
\vttab\{\vtobj{"object category"}: "Buildings", \vtattr{"attribute description"}: "Traditional East Asian-style architecture with tiled roofs", \vtknow{"knowledge description"}: "The architecture style suggests a historical or cultural significance, possibly depicting the residences or public buildings of the time."\}, \vttab\{\vtobj{"object category"}: "Tents", \vtattr{"attribute description"}: "Portable, covered shelters on the bridge", \vtknow{"knowledge description"}: "Tents may be used for shelter, storage, or as market stalls, indicating commercial or communal activities."\}, \vttab\{\vtobj{"object category"}: "Guardrails", \vtattr{"attribute description"}: "Low barriers along the edges of the bridge", \vtknow{"knowledge description"}: "Guardrails are safety features to prevent people and objects from falling off the edges of structures like bridges."\}]\}}
 \\
\bottomrule
\end{tabular}
}
\caption{An example of our generated visual tables.}
\label{table:v1_185}
\end{table*}

\begin{table*}[t]

\resizebox{\textwidth}{!}{
% \begin{tabular}[]{p{0.5in}p{4.5in}}
\begin{tabular}[]{p{1in}p{4in}}
\toprule
\multicolumn{2}{p{5in}}{
\textbf{Generated Visual Table}}\\
\midrule
\midrule
 & \includegraphics[width=0.25\linewidth]{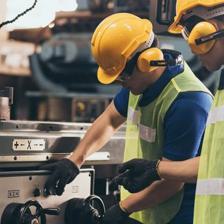} \\
\textbf{User} & \small {Based on the given image, generate the visual table that follows the following JSON format:} \textbf{......}\\
\midrule
\textbf{Our Generator} & 
{\small\{\vtred{"scene description"}: "Two workers in protective gear are operating or maintaining machinery in an industrial setting, possibly during a work shift.", "objects": [\vttab\{\vtobj{"object category"}: "Person", \vtattr{"attribute description"}: "Wearing a yellow hard hat, blue shirt, high-visibility safety vest, and black gloves; focused on a task; appears to be male.", \vtknow{"knowledge description"}: "The attire suggests the person is a worker with a role that requires safety precautions and possibly involves manual labor or operating machinery."\}, \vttab\{\vtobj{"object category"}: "Person", \vtattr{"attribute description"}: "Wearing a yellow hard hat, blue shirt, high-visibility safety vest, and black gloves; appears to be engaged in a task; appears to be male.", \vtknow{"knowledge description"}: "Similar to the first person, this individual's attire indicates a role that necessitates safety gear and manual involvement in industrial tasks."\}, \vttab\{\vtobj{"object category"}: "Safety helmet", \vtattr{"attribute description"}: "Yellow in color, worn on the head of the first person.", \vtknow{"knowledge description"}: "Safety helmets are designed to protect the head from injuries due to falling objects, impacts, or other hazards."\}, \vttab\{\vtobj{"object category"}: "Safety helmet", \vtattr{"attribute description"}: "Yellow in color, worn on the head of the second person.", \vtknow{"knowledge description"}: "This helmet serves the same protective purpose as the first, indicating a work environment where head protection is necessary."\}, \vttab\{\vtobj{"object category"}: "Protective gloves", \vtattr{"attribute description"}: "Black in color, worn by both individuals.", \vtknow{"knowledge description"}: "Protective gloves are worn to safeguard the hands from hazards such as cuts, abrasions, or exposure to materials that could cause skin irritation or disease."\}, \vttab\{\vtobj{"object category"}: "High-visibility safety vest", \vtattr{"attribute description"}: "Bright yellow-green with reflective stripes, worn by both individuals over their shirts.", \vtknow{"knowledge description"}: "High-visibility safety vests are worn to make the wearer easily visible in order to prevent accidents, especially in environments where machinery is operated or where there is vehicular traffic."\}, \vttab\{\vtobj{"object category"}: "Machinery", \vtattr{"attribute description"}: "Gray in color, with various knobs and switches, being operated or maintained by the workers.", \vtknow{"knowledge description"}: "The machinery is likely part of an industrial process, and its complexity suggests it requires skilled workers to operate or maintain it properly."\}]\}} \\
\bottomrule
\end{tabular}
}
\caption{An example of our generated visual tables.}
\label{table:160}
\end{table*}

\begin{table*}[t]

\resizebox{\textwidth}{!}{
% \begin{tabular}[]{p{0.5in}p{4.5in}}
\begin{tabular}[]{p{1in}p{4in}}
\toprule
\multicolumn{2}{p{5in}}{
\textbf{Generated Visual Table}}\\
\midrule
\midrule
 & \includegraphics[width=0.9\linewidth]{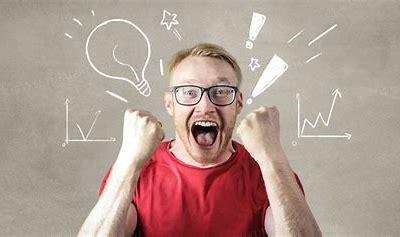} \\
\textbf{User} & \small {Based on the given image, generate the visual table that follows the following JSON format:} \\
& {\small \{\vtred{"scene description"}: "", "objects": [\{\vtobj{"object category"}: "", \vtattr{"attribute description"}: "", \vtknow{"knowledge description"}: ""\}, ......, \{\vtobj{"object category"}: "", \vtattr{"attribute description"}: "", \vtknow{"knowledge description"}: ""\}]\}} \\
\midrule
\textbf{Our Generator} & 
{\small\{\vtred{"scene description"}: "A man expressing excitement or triumph with a thought bubble containing creative ideas above his head", \newline
"objects": [
\vttab\{\vtobj{"object category"}: "Person", \vtattr{"attribute description"}: "Adult male, wearing glasses, red shirt, expressing joy with mouth open and fists raised, short hair, fair skin", \vtknow{"knowledge description"}: "The man's pose and facial expression suggest excitement, triumph, or a victorious moment. His gesture could be associated with success, achievement, or a creative breakthrough."\}, \vttab\{\vtobj{"object category"}: "Thought bubble", \vtattr{"attribute description"}: "Ideas depicted as light bulbs and stars, white and black, hand-drawn style", \vtknow{"knowledge description"}: "Thought bubbles are a graphic convention used in art to represent what a character is thinking or dreaming. The light bulbs symbolize ideas or insights, while the stars may represent success or acclaim."\}, \vttab\{\vtobj{"object category"}: "Light bulb", \vtattr{"attribute description"}: "White, six-sided, above the person's head inside the thought bubble", \vtknow{"knowledge description"}: "The light bulb is a universal symbol for an idea or a flash of inspiration. It is often used to represent creativity and the process of thinking."\}, \vttab\{\vtobj{"object category"}: "Star", \vtattr{"attribute description"}: "Yellow, five-pointed, inside the thought bubble", \vtknow{"knowledge description"}: "Stars are often used to signify excellence, success, or importance. In this context, it may be depicting a sense of achievement or celebration of an idea."\}]\}}
 \\
\bottomrule
\end{tabular}
}
\caption{An example of our generated visual tables.}
\label{table:1325}
\end{table*}

\end{document}